\documentclass[journal]{IEEEtran}
\usepackage{amsmath,amsfonts}
\usepackage{algorithmic}
\usepackage{algorithm}
\usepackage{array}
\usepackage[caption=false,font=normalsize,labelfont=sf,textfont=sf]{subfig}
\usepackage{textcomp}
\usepackage{stfloats}
\usepackage{url}
\usepackage{verbatim}
\usepackage{graphicx}
\usepackage{cite}
\usepackage{booktabs}
\usepackage{orcidlink}
\usepackage{pifont}
\usepackage{amssymb}
\usepackage{multirow}
\usepackage{threeparttable}
\usepackage{siunitx}
\usepackage{makecell}
\usepackage{colortbl}
\usepackage{xcolor}
\begin{document}
\title{Learning Video Exposure Correction:\\Large-Scale Real-World Dataset and Method}

\author{Siru Zhang\,\orcidlink{0009-0006-2764-5863}, 
Huiyuan Fu\textsuperscript{*}\,\orcidlink{0000-0002-5276-4366},~\IEEEmembership{Member,~IEEE,}
Jin Liu\,\orcidlink{0009-0008-5078-3302},

Bo Wang\,\orcidlink{0009-0002-2120-9105},
Chuanming Wang\,\orcidlink{0000-0001-6932-6226},
Chongqi Sun\,\orcidlink{0009-0004-2538-0967},
Huadong Ma\,\orcidlink{0000-0002-7199-5047},~\IEEEmembership{Fellow,~IEEE}

\thanks{
Received 30 January 2026; revised 20 July 2026; accepted 29 September 2026.
This work was supported in part by the Beijing Natural Science Foundation under Grant JQ24020, in part by the NSFC under Grant 62272059 and Grant U24B20176, and in part by the National Key R\&D Program of China under No.2023YFF0904800. \textit{(Corresponding Author: Huiyuan Fu)}.

Siru Zhang, Huiyuan Fu, Jin Liu, Bo Wang, Chuanming Wang, Chongqi Sun, and Huadong Ma are with the State Key Laboratory of Networking and Switching Technology, Beijing University of Posts and Telecommunications, Beijing 100876, China. 
(email: zhangsr@bupt.edu.cn)

}}

\markboth{IEEE Transactions on Pattern Analysis and Machine Intelligence, Vol.~1, 2026}
{Siru Zhang, 
\MakeLowercase{\textit{(et al.)}}: Learning Video Exposure Correction: Large-Scale Real-World Dataset and Method}

\IEEEpubid{0000--0000/00\$00.00~\copyright~2021 IEEE}
\IEEEoverridecommandlockouts

\IEEEpubid{\makebox[\columnwidth]{© 2026 IEEE. Personal use of this material is permitted. \hfill} \hspace{\columnsep}\makebox[\columnwidth]{ }} 

\maketitle

\begin{abstract}
Exposure correction aims to enhance visual data degraded by improper exposures, thereby improving perceptual quality. While substantial progress has been achieved in single-image exposure correction, its extension to videos remains largely underexplored.
Applying frame-wise correction methods to videos often results in flickering artifacts and degraded visual coherence.
Progress in video exposure correction has also been hindered by the lack of a large-scale paired benchmark.
To address this limitation, we introduce a real-world paired video exposure correction dataset that covers both underexposed and overexposed dynamic scenes.
The dataset comprises 36K spatially aligned frame pairs acquired using a dual-camera acquisition system equipped with a beam splitter, which enables synchronized recording of ill-exposed and well-exposed videos.
Based on this dataset, we develop an end-to-end dual-stream Retinex framework that jointly handles underexposure and overexposure through exposure-aware temporal modeling.
To mitigate exposure-induced interference during temporal alignment, our framework aggregates the amplitude spectra of neighboring frames into a shared exposure representation while retaining phase-derived structural information for cross-frame correspondence.
The framework further reinterprets residual inter-frame discrepancies as pseudo-gradient cues for inverted attention, enabling temporal fusion to preserve complementary information while suppressing unreliable content.
Finally, a learned restoration module combines the disentangled reflectance and dual-stream illumination components to produce temporally consistent exposure-corrected frames.
The extensive experiments based on various evaluations and user studies demonstrate the significance of our dataset and the effectiveness of our method.
Our dataset and code are available at \url{https://github.com/zhangisland/VECNetPlus/}.
\end{abstract}


\section{Introduction}

\IEEEPARstart{I}{n} real-world scenarios, visual data are frequently captured under complex illumination conditions, resulting in underexposure or overexposure and thus degraded visual quality.
Correcting improperly exposed images and videos is a long-standing problem in multimedia and computer vision, which is critical for photography and mobile imaging applications.
Despite recent progress, exposure correction remains non-trivial, as real-world scenes often exhibit highly variable illumination conditions and diverse capture settings.

In particular, effective exposure correction requires the ability to handle both underexposed and overexposed inputs within a unified framework simultaneously. 
However, underexposed frames require the recovery of missing visual details through amplification, whereas overexposed frames demand the suppression of saturated content. 
Such intrinsically opposite restoration objectives complicate stable optimization. 
Moreover, the difficulty is further amplified when extending exposure correction from images to videos.
Specifically, in dynamic scenes, complex motion patterns and exposure variations induce non-stationary temporal statistics, and thus make naive frame-wise processing unstable.
Consequently, effective video exposure correction requires accurate frame-level exposure modeling and mechanisms to ensure temporal coherence under varying conditions.

\IEEEpubidadjcol

\begin{figure}[!tp]
    \centering 
    \includegraphics[width=1.0\linewidth]{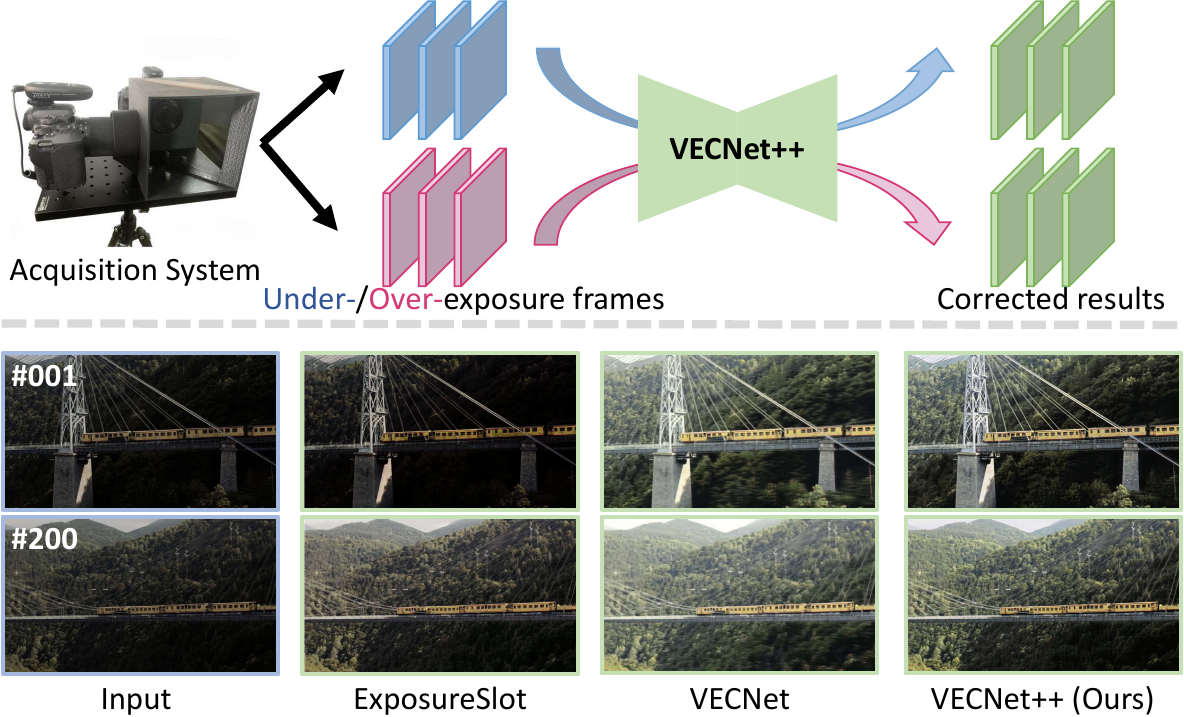}
    \caption{Top: Overview of the benchmark for the proposed video exposure correction task. Bottom: Visual comparison with the image-based method ExposureSlot~\cite{jungExposureslotExposurecentric2025a} and the video-based methods VECNet~\cite{liuLearningExposure2024} and the proposed VECNet++. Image-based method exhibits temporally inconsistent exposure, while VECNet++ achieves temporally consistent correction with improved visual clarity compared to VECNet.}
    \label{fig:intro}
    \vspace{-1em}
\end{figure}

However, existing datasets and methods do not fully address these challenges.
In recent years, driven by several representative paired image exposure datasets, such as SICE~\cite{caiLearningDeep2018}, MSEC~\cite{afifiLearningMultiScale2021}, and LCDP~\cite{wangLocalColor2022}, a number of learning-based image exposure correction methods~\cite{huangDeepFourierBased2022,huangExposureConsistencyRepresentation2022,wangLocalColor2022,huangLearningSample2023,wangDecouplingandAggregatingImage2023,yaoGeneralizedLightness2023} have been proposed.
Despite recent progress, these methods are designed for single images and do not incorporate inter-frame modeling for temporal coherence. 
When directly applied to videos in a frame-by-frame manner, they often suffer from severe visual flickering and exposure inconsistency, as illustrated in Fig.~\ref{fig:intro}.

There are several related video-based approaches from low-light video enhancement~\cite{weiDeepRetinex2018,wangSeeingDynamic2021,fuDancingDark2023} and High Dynamic Range (HDR) video reconstruction~\cite{chenHDRVideo2021,shuRealworldHDR2024}.
However, these methods optimize for low-light enhancement or HDR reconstruction, which fundamentally differ from the objectives of single-exposure video correction.
Low-light video methods primarily target visibility improvement for underexposed inputs, whereas HDR reconstruction typically relies on multi-exposure or exposure-bracketed inputs with radiometric alignment to recover high dynamic range representations.
These differences limit the suitability of such methods to general single-exposure video exposure correction.

Motivated by these challenges, we present the first study that focuses on real-world video exposure correction. 
To the best of our knowledge, real-world video exposure correction has not been adequately addressed,
largely due to the lack of a dedicated benchmark dataset and a systematic solution. Therefore, we jointly advance dataset construction and method design to fill this gap.

\noindent \textbf{(1) Benchmark Dataset.}
Existing research on video exposure correction is limited by the lack of a large-scale, high-quality benchmark. 
To address this limitation, we construct a high-quality paired video dataset for exposure correction, covering both underexposed and overexposed conditions in dynamic real-world scenes.
Collecting such data is non-trivial, as it requires acquiring ill-exposed and well-exposed video pairs of the same scene with consistent camera motion and accurate spatial-temporal alignment.
To facilitate the dataset collection, we design a dual-camera acquisition system equipped with a beam splitter for parallax-free capture under different exposure settings.
The captured reference videos are carefully post-processed to provide reliable ground-truth supervision.
Based on the system, we construct a dataset comprising 119 high-quality video pairs, termed DIME (\textbf{D}ynamic scenes \textbf{I}n \textbf{M}ultiple \textbf{E}xposure)~\cite{liuLearningExposure2024}, featuring diverse scenes and complex motion patterns with precise spatial-temporal alignment.

\noindent \textbf{2) Video Exposure Correction Method.}
Based on our dataset, we propose an end-to-end Video Exposure Correction Network (VECNet) to restore videos with improper exposure.
Unlike frame-wise exposure correction, video exposure correction requires learning the temporally stable exposure representation.
To ensure reliable temporal representations, VECNet aggregates the Fourier amplitudes of neighboring frames into a shared representation and recombines it with their phase components for subsequent alignment. 
The aligned temporal features are then restored using dual-stream illumination modeling to handle underexposed and overexposed regions within a unified framework.
This design enables the model to exploit temporal information while reducing exposure-induced inconsistency across frames.

More importantly, this manuscript presents a substantial extension and improvement over our earlier ACM MM 2024 Oral work~\cite{liuLearningExposure2024}.
Compared with the conference version, this journal version introduces significant advancements in the following aspects:
\begin{itemize}
    \item \textbf{Dataset Extension.}
    We substantially expand the original DIME dataset to a larger-scale version (DIMEPlus) from 119 to 192 video sequences, and from 21,951 to 36,099 frames in total.
    The extended dataset DIMEPlus contains more diverse scenes and motions while preserving precise spatial-temporal alignment. Specifically, compared with DIME, DIMEPlus achieves a lower LOE score (14.58 vs.\ 14.91), indicating improved exposure consistency, and exhibits increased motion diversity, reflected by a higher average optical flow magnitude (1.45 vs.\ 1.34).
    \item \textbf{Methodological Advancement.}
    We further develop VECNet++ to explicitly address the limitations of the prior VECNet~\cite{liuLearningExposure2024} in handling pixel-level misalignment and inter-frame semantic inconsistency under complex motion.
    The key design of VECNet++ is to incorporate exposure variations directly into temporal correspondence estimation and feature fusion. This reduces exposure-induced interference during alignment and allows residual inter-frame discrepancies to guide which supporting features are retained during fusion.
    In contrast to the original version, VECNet++ achieves advanced performance by (i) revisiting dual-stream illumination estimation with a deterministic pyramid convolutional design that preserves temporal coherence while reducing model complexity, (ii) adding an optical-flow-based multi-frame pre-alignment to complement frequency-domain alignment, 
    and (iii) introducing a novel Exposure Adaptive Fusion (EAF) module by exploiting the exposure-invariant and exposure-varying features to adaptively fuse reliable details across frames while avoiding exposure-induced artifacts.
    \item \textbf{Comprehensive Experimental Evaluation.}
    We conduct extensive experiments on DIMEPlus and additional real-world video datasets, i.e., Real-HDRV~\cite{shuRealworldHDR2024}, LLIV-Phone~\cite{liLowLightImage2022}, and our newly collected ProCaptureVideo, demonstrating the effectiveness and generalization of the proposed dataset and methods.
\end{itemize}

Our contributions can be summarized as follows:
\begin{enumerate}
    \item We establish a high-quality paired video exposure correction benchmark for dynamic real-world scenes, termed DIMEPlus, which contains 192 sequences (36,099 frames) with precise spatial-temporal alignment and diverse exposure conditions, enabling systematic evaluation of video exposure correction.
    \item We propose an end-to-end video exposure correction framework that couples dual-stream illumination modeling with multi-frame temporal alignment. We further develop VECNet++ to jointly improve exposure-aware correspondence, reliable temporal fusion, and illumination-guided restoration for enhanced temporal robustness and visual fidelity.
    \item Comprehensive evaluations on multiple real-world datasets that establish a reproducible benchmark and demonstrate the effectiveness of the proposed approach.
\end{enumerate}

\section{Related Works}
\label{sec:related_work}
Video exposure correction is currently underexplored, and in this section, we provide related works on datasets, image-level exposure correction, and video-level restoration.

\subsection{Datasets}
\label{subsec:dataset}
Existing image-level exposure correction datasets, such as SICE~\cite{caiLearningDeep2018}, MSEC~\cite{afifiLearningMultiScale2021}, and LCDP~\cite{wangLocalColor2022}, provide paired images for training and evaluation. 
While these datasets have been instrumental for single-image methods, they focus solely on static images and do not capture temporal dynamics necessary for video exposure correction. 
As summarized in Tab.~\ref{tab:comparison_datasets}, existing real-world video datasets primarily focus on either underexposed scenarios or exposure stacks for HDR reconstruction, while large-scale paired datasets covering both underexposed and overexposed videos in dynamic scenes remain scarce. 
It motivates us to construct a large-scale video exposure correction dataset.

Video datasets have been developed for related tasks, including low-light enhancement and HDR reconstruction. 
However, these tasks are defined under different problem settings, in terms of both inputs and target objectives.
Representative low-light video datasets include DRV~\cite{chenSeeingMotion2019}, SDSD~\cite{wangSeeingDynamic2021}, DID~\cite{fuDancingDark2023}, and R-DID~\cite{wangRethinkingLowLight2025a}.
These datasets only contain underexposed samples, and the absence of overexposed samples results in insufficient data for the exposure correction task.
HDR video datasets, such as DeepHDRVideo~\cite{chenHDRVideo2021}, LDR-HDR~\cite{yueHDRVideo2023}, and Real-HDRV~\cite{shuRealworldHDR2024}, are designed for multi-exposure fusion and radiance recovery. 
These datasets generate HDR labels by fusing LDR images, which are intended for HDR reconstruction rather than display-ready, well-exposed LDR videos, limiting their direct applicability for video exposure correction.

In summary, existing datasets are either limited to single images or designed for related tasks such as low-light enhancement and HDR reconstruction, leaving video exposure correction without dedicated benchmarks.

\begin{table}[t]
\centering
\begin{threeparttable}
\caption{Comparisons between existing real-world video datasets and our proposed DIME and DIMEPlus.}
\label{tab:comparison_datasets}
\setlength{\tabcolsep}{3pt}
\begin{tabular}{l@{}ccrcc}
\toprule
Dataset & Dynamic&\text{\#Sequences} & \text{\#Pairs} & Resolution & Exposure \\
\midrule
DRV~\cite{chenSeeingMotion2019}&\ding{55} &202&202& 3672$\times$5496 & U \\ 
SDSD~\cite{wangSeeingDynamic2021}&\ding{51} &150&37,500& 1920$\times$1080 & U \\ 
DID~\cite{fuDancingDark2023}&\ding{51} &413&41,038& 2560$\times$1440 & U \\
R-DID~\cite{wangRethinkingLowLight2025a}&\ding{51} &78&30,305& 1280$\times$720 & U \\
\midrule
DeepHDRVideo~\cite{chenHDRVideo2021}&\ding{51} &281&$\sim$1,800& 4096$\times$2168 & Stack \\
Real-HDRV~\cite{shuRealworldHDR2024}&\ding{51} &500&4,000& 1500$\times$1000 & Stack \\
\midrule
DIME~\cite{liuLearningExposure2024} (Ours)&\ding{51}& 119 & 21,951 & 1920$\times$1080 & U+O \\ 
DIMEPlus (Ours)&\ding{51}& 192 & 36,099 & 1920$\times$1080 & U+O \\
\bottomrule
\end{tabular}
\begin{tablenotes}
    \item ``U'' and ``O'' indicate underexposed and overexposed pairs.
    \item ``Stack'' indicates exposure stacks (multi-exposure frames of the same scene).
    \item ``\#Pairs'' counts only pairs with GT labels for training or evaluation.
\end{tablenotes}
\end{threeparttable}
\vspace{-1.5em}
\end{table}

\subsection{Image-level Enhancement Methods}

\subsubsection{Low-light Enhancement}
\label{subsubsec:lowlight}
Low-light enhancement is closely related to exposure correction, as both aim to improve visual quality under ill-exposed conditions.
However, existing low-light enhancement methods are designed to recover the underexposed content, rather than addressing diverse exposure conditions in a unified manner.
Recent low-light enhancement approaches are developed based on Retinex theory~\cite{wangSeeingDynamic2021,liLearningEnhance2022,fuDancingDark2023,caiRetinexformerOnestage2023,yanHVINew2025,liaoGTMeanLoss2025,liangAestheticsGuidedLowLight2025,huangQuadPriorMultidimension2025}. 
For example, SDSDNet~\cite{wangSeeingDynamic2021} employs feature pyramids to aggregate multi-frame information and reconstruct enhanced frames by fusing progressively estimated illumination and noise components.
LAN~\cite{fuDancingDark2023} decomposes images into illumination and reflectance components and adaptively adjusts illumination through an autoencoder-based latent representation.
Retinexformer~\cite{caiRetinexformerOnestage2023} proposes a one-stage Retinex-based framework to restore the corruptions in the light-up process.
HVI-CIDNet~\cite{yanHVINew2025} rethinks the low-light image enhancement from the perspective of the HVI color space.
Recently, GTMeanLoss~\cite{liaoGTMeanLoss2025} addresses brightness bias in low-light enhancement by introducing a probabilistic loss to model image mean intensity.
Despite their effectiveness in enhancing visibility under low-light conditions, these methods are inherently tailored to underexposed inputs and do not explicitly consider overexposed content.
This restricts their applicability to video exposure correction involving diverse exposure conditions.

\subsubsection{Image Exposure Correction}
Previous traditional image correction methods~\cite{Pizer1987HE,Prizer1990CLAHE,guanImageEnhancement2009}, e.g., histogram equalization and gamma correction, mainly rely on manipulating basic color and intensity parameters.
Such approaches are typically heuristic-driven and lack the ability to model complex, scene-dependent degradations.
With the development of deep learning, there are emerging learning-based exposure correction methods.
Most existing approaches are developed under the single-image setting and can be broadly categorized according to their illumination modeling strategies.

(i) Single-stream illumination modeling.
These methods directly learn global or regional exposure adjustment functions without explicitly disentangling illumination components.
Some approaches estimate image enhancement parameters or filter arguments in a data-driven manner~\cite{cuiYouOnly2022,keHarmonizerLearning2022}.
DRBN-ENC~\cite{huangExposureNormalization2022} introduces instance normalization to reduce exposure variation, while ECLNet~\cite{huangExposureConsistencyRepresentation2022} exploits exposure-consistent features through a bilateral activation mechanism.
FECNet~\cite{huangDeepFourierBased2022} leverages Fourier representations to mitigate exposure imbalance by modeling image frequency components.
LACT~\cite{baekLuminanceawareColor2023b} models relative exposure relationships to guide luminance adjustment.
More recent methods incorporate region-aware processing or frequency-aware representations to improve global exposure consistency~\cite{liuRegionAwareExposure2024,zhou4KResolutionPhoto2024,liRealTimeExposure2024,jungExposureslotExposurecentric2025a}.
Despite their effectiveness on single images, these methods typically assume a unified exposure adjustment process and do not explicitly model the distinct characteristics of underexposure and overexposure.

(ii) Dual-stream illumination modeling.
To explicitly handle the underexposure and overexposure conditions, dual-stream illumination strategies have been explored.
The concept can be traced back to the non-learning-based DUAL framework~\cite{zhangDualIllumination2019}, which separately processes underexposed and overexposed regions.
This idea has been extended to deep learning paradigms, such as LCDP~\cite{wangLocalColor2022}, which learns local color distributions under dual-stream illumination branches.
CSEC~\cite{liColorShift2024} utilizes the dual-stream illumination branches to estimate color distribution shifts between the two ill-exposed regions and learns how to correct the colors in extremely exposed regions.
While dual-stream illumination modeling improves robustness to diverse exposure patterns at the image level, existing methods are limited to static inputs and do not consider temporal coherence when applied to videos.

\subsection{Video-level Enhancement Methods}
Extending image-level enhancement methods to videos is challenging, as naive frame-wise processing often leads to temporal artifacts such as flickering and inconsistent illumination across frames.

\subsubsection{Low-light and HDR Video Enhancement}
Beyond image-level enhancement methods, there are several related video enhancement methods for low-light enhancement and HDR reconstruction.
As discussed in Sec.~\ref{subsubsec:lowlight}, low-light video enhancement methods~\cite{wangSeeingDynamic2021,fuDancingDark2023} primarily focus on improving visibility in underexposed videos and do not address overexposed content or mixed exposure scenarios.
Meanwhile, representative HDR video reconstruction methods, such as CoarseNet~\cite{chenHDRVideo2021} and Shu et al.~\cite{shuRealworldHDR2024}, typically rely on multi-exposure inputs or exposure-bracketed sequences with strict radiometric alignment.
In contrast, video exposure correction targets single-exposure LDR videos and aims to produce temporally coherent, display-ready results without explicit radiance supervision.
As a result, existing low-light and HDR video methods are not directly applicable to video exposure correction under diverse exposure conditions.

\subsubsection{General Video Restoration}
General video restoration methods aim to address degradations, including noise, blur, and compression artifacts, by leveraging temporal correlations across frames.
These methods typically exploit neighboring frames using strategies such as explicit temporal alignment~\cite{wangEDVRVideo2019,zhouRevisitingTemporal2022,liSimpleBaseline2023}, recurrent architectures for sequential modeling~\cite{luongDesigningInterpretable2021,liangRecurrentVideo2022}, and attention mechanisms for feature aggregation~\cite{liangVRTVideo2024,niHazyLowQuality2025}. 

While effective when inter-frame exposure levels are consistent, these approaches can fail when exposure varies significantly between frames. 
In such cases, directly fusing multiple frames may amplify inconsistencies and propagate unreliable details. 
Inspired by detection tasks using inverted or complementary attention~\cite{huangImprovingObject2020}, we introduce an exposure-aware attention mechanism that selectively propagates reliable features and suppresses exposure-induced artifacts. 
This design complements existing temporal fusion strategies and addresses the challenges of dynamic exposure variations in video exposure correction.

\section{Our Proposed DIMEPlus Dataset}
\label{sec:dataset}
Video exposure correction requires large-scale paired video data captured under diverse exposure conditions.
In particular, this task demands real-world videos containing under- and over-exposed inputs, paired with corresponding well-exposed ground-truth videos.
As discussed in Sec.~\ref{sec:related_work}, such datasets are currently not publicly available for video exposure correction. 
To address this gap, we construct a real-world dataset covering diverse dynamic scenes and varying exposure conditions.

\begin{figure}[!tp]
    \centering
    \includegraphics[width=.85\linewidth]{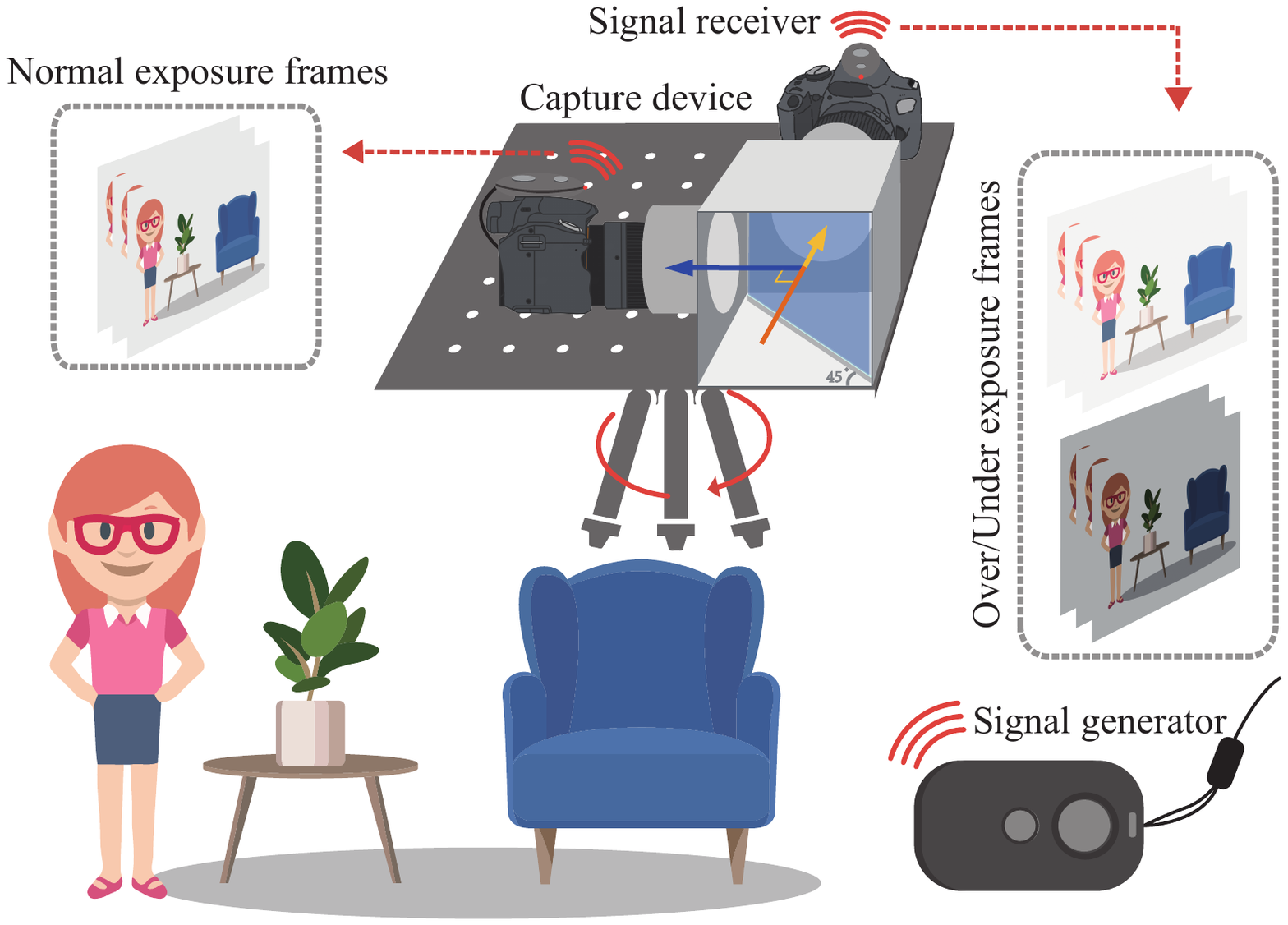}
    \caption{We built an optical system to capture the normal and over-/under- exposure video pairs.}
    \label{fig:system}
\end{figure}

\subsection{Hardware Design}
Capturing paired multi-exposure videos in dynamic scenes is technically challenging due to various motion patterns and camera synchronization requirements.
While single-exposure adjustable DSLRs~\cite{fuDancingDark2023} or electric slide rails~\cite{wangSeeingDynamic2021} are effective for static scenes, they are impractical for capturing diverse dynamic scenes. To overcome this, we employ two synchronized cameras with different apertures and ISO settings. 
To keep the cameras in sync, we use an infrared remote control to trigger two camera-mounted receivers for capturing simultaneously. 

To prevent parallax and ensure consistent motion trajectories, we adopt a beam splitter to divide the incident light equally into both lenses, as illustrated in Fig.~\ref{fig:system}. The cameras and beam splitter are mounted on a custom 3D-printed opaque box fixed to an optical breadboard and a tripod to ensure stability. This design allows both cameras to capture the same scene from identical viewpoints under different exposure conditions.

\subsection{Data Collection}
We use two Canon EOS R10 cameras to capture paired videos. 
Specifically, to minimize the loss of detail in shadows and highlights, we set one camera to automatic exposure mode and record video in Canon Log, which retains more visual information throughout the dynamic range. 
Then we enroll professional photographers to produce high-quality, well-exposed reference videos using DaVinci Resolve Studio. 
To maintain consistent exposure correction across the reference video, the adjustment process is restricted to basic exposure-related corrections, such as white balance and tone-mapping adjustment, and is applied consistently at the video level.
The other camera captures under-/overexposed videos in standard sRGB format. All the other settings are kept at default values to simulate real capture scenarios. 
Each dynamic scene is recorded as a 10-second paired video at 50 fps.

\begin{figure*}[!t]
    \centering
    \includegraphics[width=1.0\linewidth]{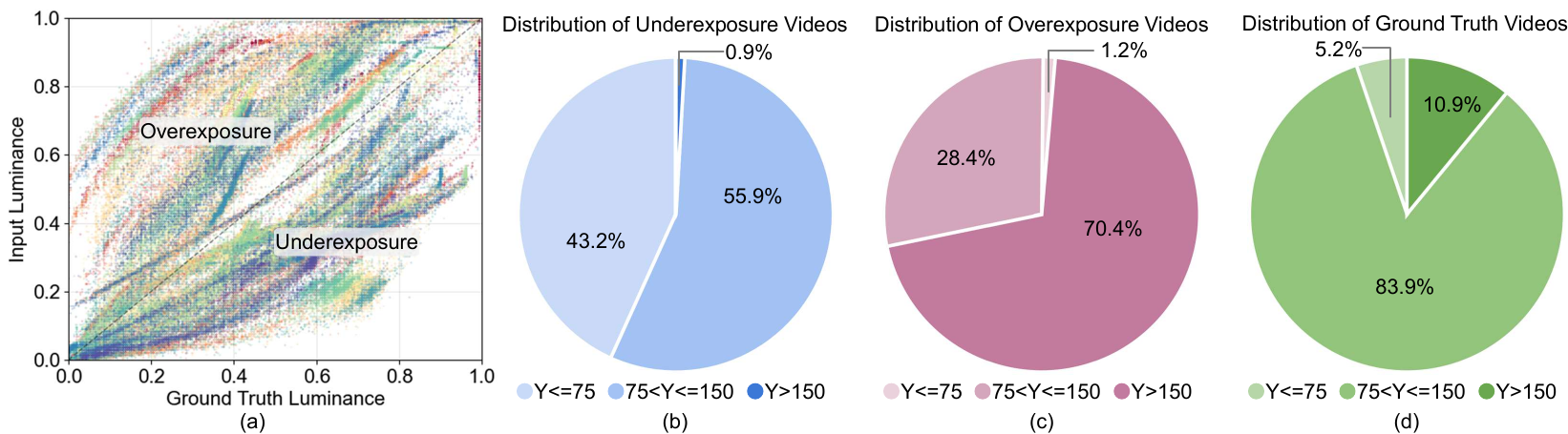}
    \caption{Statistics of the DIMEPlus dataset. (a) The input-ground truth brightness mapping curve statistics. Each cluster corresponds to a distinct video pair, demonstrating the broad coverage of our dataset. (b–d) Luminance ($Y$) distributions for underexposed, overexposed, and ground-truth videos, respectively.}
    \label{fig:dataset_analysis}
\end{figure*}

\begin{figure}[h]
    \centering
    \includegraphics[width=1.0\linewidth]{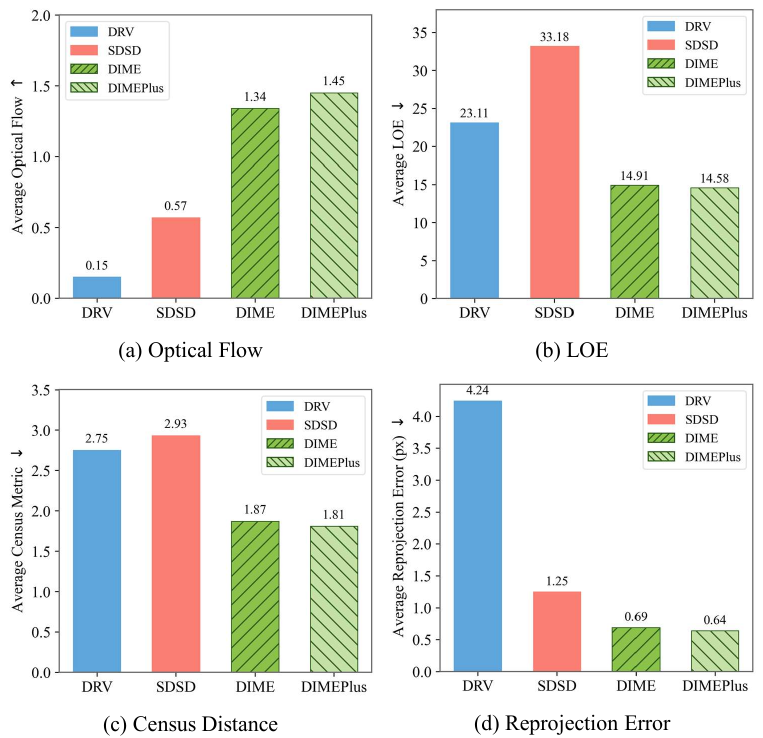}
    \caption{Comparison of average optical flow magnitude, LOE, Census distance, and reprojection error across datasets.}
    \label{fig:dataset_metrics}
\end{figure}

\subsection{Data Processing}
Due to minor inaccuracies in the manual assembly of the hardware system, a slight spatial misalignment remains between the paired videos.
To address this issue, we adopt a two-stage alignment strategy. First, we estimate a homography matrix between the under-/overexposed frame and the corresponding normally exposed frame using the matched SIFT~\cite{loweObjectRecognition1999} keypoints. 
In this way, we can roughly crop the frames to corresponding regions. Then we utilize a traditional flow estimation algorithm, DeepFlow~\cite{weinzaepfelDeepFlowLarge2013}, to perform pixel-wise alignment. Finally, we use center cropping to remove alignment artifacts around boundaries, producing precise spatially aligned frames. 
Note that the alignment transformation is applied only to the normally exposed frames, while the ill-exposed inputs remain unchanged to preserve their original capture characteristics.
Additionally, we use optical flow~\cite{farnebackTwoframeMotion2003}, lightness-order-error (LOE)~\cite{wangNaturalnessPreserved2013}, Census distance~\cite{rukundo20194x4censustransform}, and reprojection error to quantitatively evaluate motion magnitude and alignment performance.
Fig.~\ref{fig:dataset_metrics}(a) reports the optical-flow magnitude, which characterizes the motion level across different datasets.
Figs.~\ref{fig:dataset_metrics}(b)-(d) report the LOE, Census distance, and reprojection error results, respectively. 
LOE measures the consistency of relative lightness ordering, while Census distance compares local binary structural descriptors at corresponding spatial locations and is relatively robust to monotonic photometric variations. Reprojection error measures the residual pixel displacement between matched points in the paired frames.
These results show that the paired frames in our dataset exhibit larger motion and lower residual misalignment than those in existing low-light video enhancement datasets, including DRV~\cite{chenSeeingMotion2019} and SDSD~\cite{wangSeeingDynamic2021}, demonstrating their improved suitability for dynamic real-world scenes.

\begin{figure}[t]
    \centering
    \includegraphics[width=1.0\linewidth]{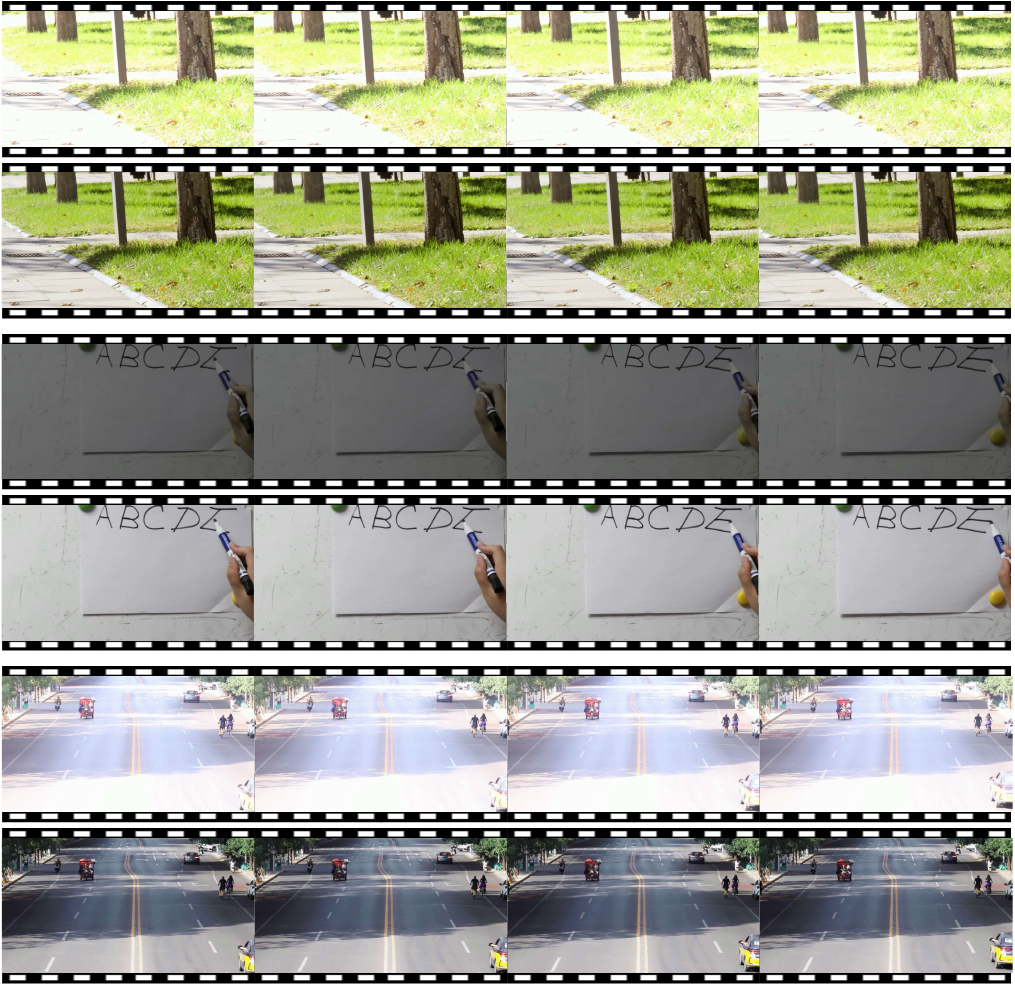}
    \caption{Representative video samples from our dataset. Frames are shown every five frames to better visualize motion. The dataset covers under-/over-exposures, indoor/outdoor scenes, and camera/object motions.}
    \label{fig:data_samples}
\end{figure}

\subsection{Dataset Statistics}
The collected videos are segmented into continuous, non-overlapping clips of 200 frames each, yielding a total of 192 video clips and 36,099 frames at 1920$\times$1080.
Fig.~\ref{fig:dataset_analysis} presents a statistical analysis of exposure conditions in the proposed DIMEPlus dataset. 
Following previous works~\cite{wangLocalColor2022,liuLearningExposure2024}, the luminance mapping between input and ground-truth videos is visualized in Fig.~\ref{fig:dataset_analysis} (a). 
Each cluster corresponds to an individual video sequence, demonstrating that DIMEPlus spans a broad spectrum of underexposed and overexposed conditions rather than being concentrated around well-exposed regions.
To further quantify the exposure characteristics, a detailed analysis is conducted for underexposed, overexposed, and ground-truth videos. Fig.~\ref{fig:dataset_analysis} (b-d) reports the exposure composition of these three categories based on luminance levels. The results indicate that the dataset maintains a balanced coverage across diverse exposure conditions, enabling comprehensive evaluation under realistic lighting variations.

Three representative video pairs are illustrated in Fig.~\ref{fig:data_samples}. From top to bottom, the examples include overexposed and underexposed scenarios, outdoor and indoor environments, as well as camera motion and object motion. 
These statistics indicate that the dataset covers diverse scene types and motion patterns, enabling evaluation across a wide range of real-world conditions.

\section{Methodology}
\subsection{Overview}
The Video Exposure Correction (VEC) task aims to learn a mapping function $\mathcal{\varGamma}$ that transforms consecutive improperly exposed video frames into visually consistent and properly exposed outputs. 
Formally, given $2N+1$ consecutive 8-bit sRGB frames $\mathcal{I}_{[t-N:t+N]}$, the goal is to estimate the corrected reference frame $\mathcal{O}_t$ as:
\begin{equation}
    \label{eq:task_formulation}
    \mathcal{O}_t = \mathcal{\varGamma}(\mathcal{I}_{[t-N:t+N]}),
\end{equation}
where the center frame $\mathcal{I}_t$ serves as the reference, and the remaining frames provide temporal support.
Despite this straightforward formulation, VEC remains challenging due to severe exposure variations across frames and temporal misalignment that jointly hinder consistent reflectance recovery.

\begin{figure*}[!t]
    \centering
    \includegraphics[width=1.0\linewidth]{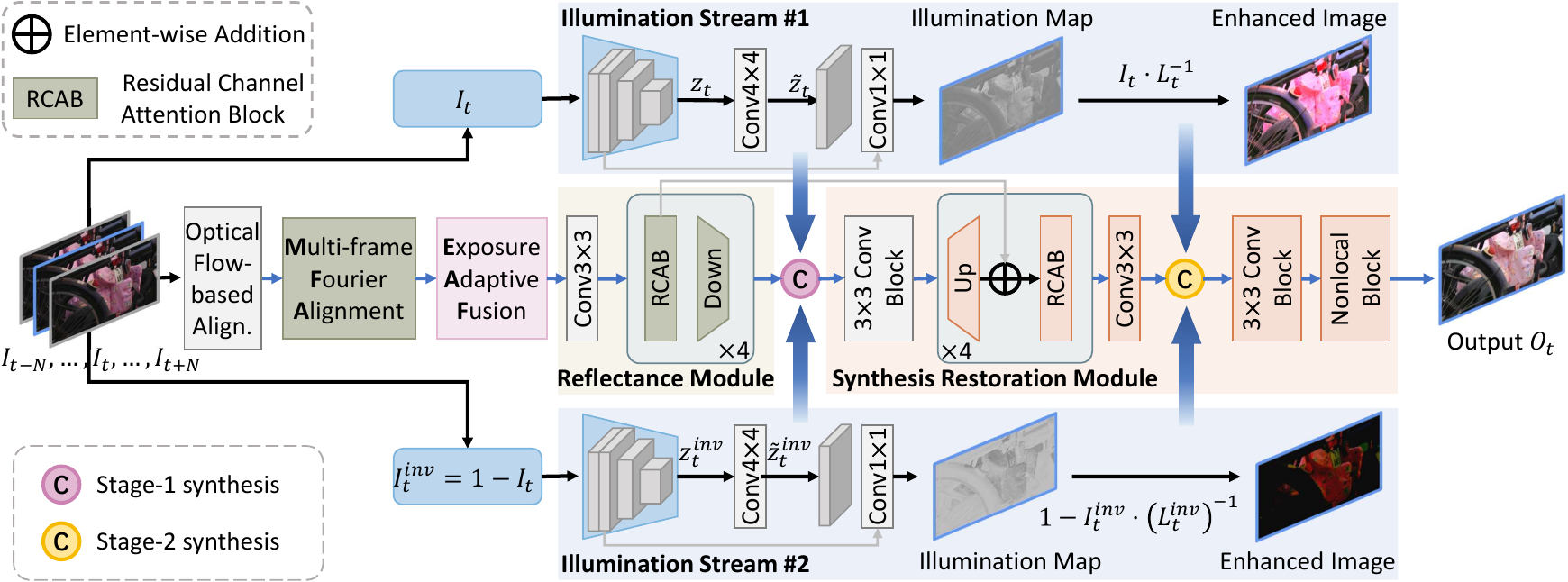}
    \caption{Overview of our framework VECNet++. It contains four modules, including Multi-frame Fourier Alignment (MFA), Exposure Adaptive Fusion (EAF), Dual-stream Illumination Construction (DIC), and Two-stage Synthesis Restoration (TSR).}
    \label{fig:model_arch}
    \vspace{-1em}
\end{figure*}

Motivated by the illumination-reflectance decomposition principle commonly adopted in Retinex-based methods, we observe that directly extending single-frame Retinex models to video often leads to temporal inconsistency and unstable illumination estimation under large exposure fluctuations.
To address these challenges in a principled manner, we design VECNet++, as illustrated in Fig.~\ref{fig:model_arch}.

VECNet++ consists of four core components: 
(i) a Multi-frame Fourier Alignment (MFA) module (Sec.~\ref{subsec:multiframe_fourier_alignment}) to establish exposure-robust temporal correspondence, 
(ii) an Exposure Adaptive Fusion (EAF) module (Sec.~\ref{subsec:exposure_adaptive_fusion}) to reconcile residual misalignment and exposure discrepancies,
(iii) a Dual-stream Illumination Construction (DIC) unit (Sec.~\ref{subsec:dual_illu_construction}) to model underexposed and overexposed regions in a symmetric and disentangled manner, and
(iv) a Two-stage Synthesis Restoration (TSR) unit (Sec.~\ref{subsec:twostage_synthesis_restoration}) to progressively synthesize illumination and reflectance cues into the final output.

Specifically, to enable consistent reflectance learning across frames, the MFA module first aligns supporting frames to the reference frame while suppressing exposure-induced interference.
The EAF module further mitigates residual misalignment by performing ``exposure-invariant'' normalization and inverted attention-guided fusion, compensating for missing details caused by large exposure discrepancies adaptively. Subsequently, the DIC unit constructs dual illumination representations to handle underexposed and overexposed regions symmetrically. Finally, the TSR unit synthesizes the illumination and reflectance cues at both feature and image levels to produce the final exposure-corrected frame $\mathcal{O}_t$.

\begin{figure}[!t]
    \centering
    \includegraphics[width=1.0\linewidth]{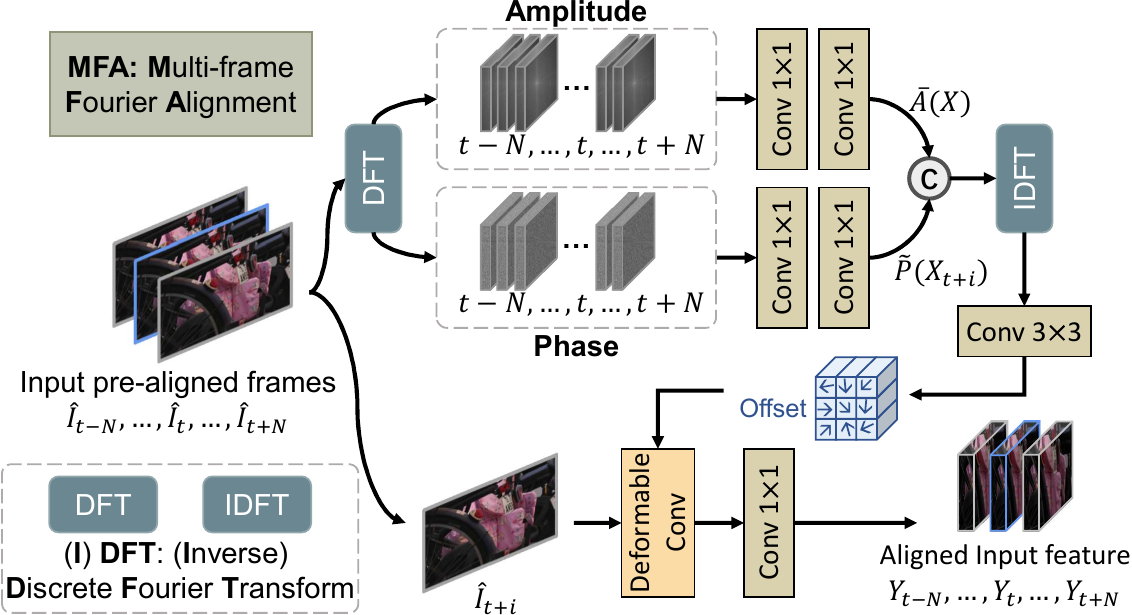}
    \caption{Details of the Multi-frame Fourier Alignment module. We apply the discrete Fourier transformation to map the frames from pixel space to Fourier space, then adopt the deformable convolution for feature-level alignment.}
    \label{fig:MFA_module}
    \vspace{-1em}
\end{figure}

\begin{figure*}[!t]
    \centering
    \includegraphics[width=1.0\linewidth]{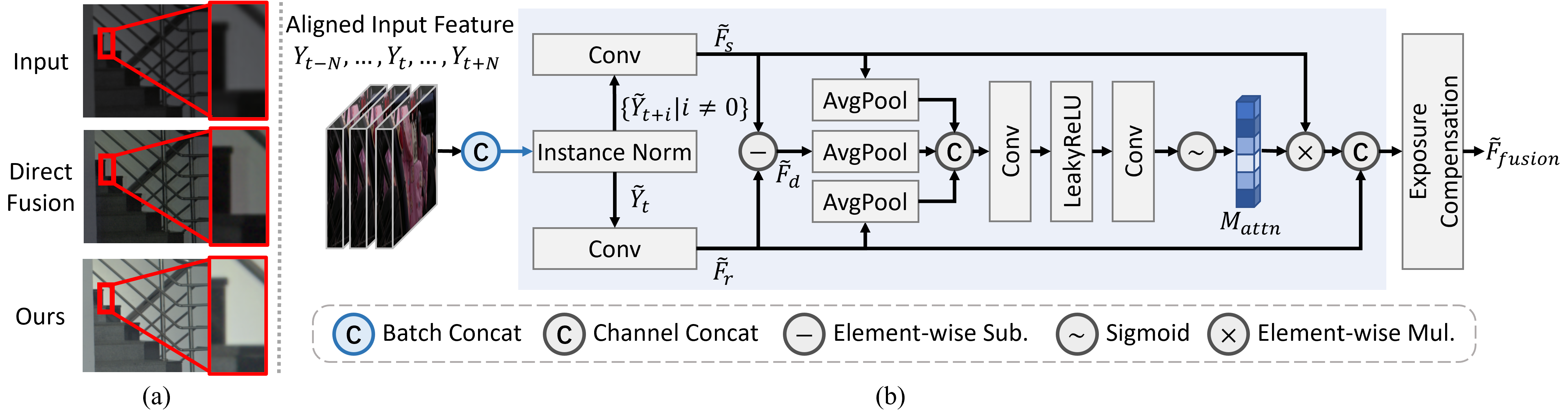}
    \caption{(a) Direct fusion of stacked multi-frame inputs produces overlay stair artifacts, whereas the proposed fusion strategy (Ours) effectively suppresses them. (b) Overview of the Exposure Adaptive Fusion module. The difference between supporting frames ${Y_{t-N},\dots,Y_{t+N}}\setminus Y_t$ and the reference frame $Y_t$ is used as a pseudo-gradient to compute the inverted attention map, guiding selective fusion across frames.}    
    \label{fig:EAF_module}
    \vspace{-1em}
\end{figure*}

\subsection{Multi-frame Fourier Alignment}
\label{subsec:multiframe_fourier_alignment}
A key challenge in video exposure correction lies in handling spatial misalignment and intensity inconsistencies across frames, particularly under severe exposure variations.
While multi-frame inputs provide temporal cues, naive fusion of misaligned frames introduces blur and ghosting artifacts~\cite{tianTDANTemporallyDeformable2020}. Existing video restoration approaches rely on feature-level warping via deformable convolutions or pyramid alignment~\cite{wangEDVRVideo2019,yueRealRawVSRRealWorld2022}, but these strategies are susceptible to varied exposure across frames, which hinders alignment estimation.

\noindent\textbf{Optical-flow-based pre-alignment.} 
To roughly reduce large spatial displacements and provide a coarse geometric prior, we apply an optical-flow-based pre-alignment as an initial step.
Specifically, we estimate the flow between each supporting frame $\mathcal{I}_{t+i}$ and the reference frame $\mathcal{I}_t$ using RAFT~\cite{teedRAFTRecurrent2020}, and warp the supporting frames accordingly:
\begin{equation}
    \hat{\mathcal{I}}_{t+i} = \text{Warp}(\mathcal{I}_{t+i}, \text{Flow}(\mathcal{I}_{t+i}, \mathcal{I}_t)).
\end{equation}
These pre-aligned frames $\hat{\mathcal{I}}_{t+i}$ provide coarse pixel-level alignment for subsequent multi-frame processing.

\noindent\textbf{Fourier-based multi-frame alignment.} After pre-alignment, we perform frequency-domain alignment to suppress exposure variations while preserving structural information.

Following prior observations in Fourier-based exposure modeling~\cite{huangDeepFourierBased2022}, exposure variations predominantly affect the amplitude spectrum, while structural information remains relatively stable in the phase component.
Leveraging this property, we design a Multi-frame Fourier Alignment (MFA) module to suppress exposure-induced interference during temporal alignment.

Specifically, given a channel in a single frame $x\in \mathbb{R}^{H\times W}$, the Fourier transform $\mathcal{F}$ converts $x$ to a complex component $X$ in Fourier space, which can be written as:
\begin{equation}
        \mathcal{F}(x)=\frac1{\sqrt{HW}}\sum_{h=0}^{H-1}\sum_{w=0}^{W-1}x(h,w)e^{-j2\pi(\frac hHu+\frac wWv)}.
\end{equation}
Then, the amplitude $\mathcal{A}(X)$ and phase $\mathcal{P}(X)$ components can be divided from the complex component $X$:
\begin{equation}
        \mathcal{A}(X)=\sqrt{R^{2}(X)+I^{2}(X)},\quad \mathcal{P}(X)=arctan\frac{I(X)}{R(X)},
\end{equation}
where $R$ and $I$ denote the real and imaginary parts of the complex $X$. We also apply the same strategy to each channel of input frames and then get consecutive amplitude components $\mathcal{A}(X_{[t-N:t+N]})$ and phase components $\mathcal{P}(X_{[t-N:t+N]})$. Since the amplitude component responds more to the lightness, we obtain $\bar{\mathcal{A}}(X)$ by aggregating multi-frame amplitude to achieve a unified exposure representation:
\begin{equation}
        \bar{\mathcal{A}}(X)=\mathcal{H}([\mathcal{A}(X_{t-N}), ..., \mathcal{A}(X_{t+N})]),
\end{equation}
where $\mathcal{H}$ denotes an aggregation network composed of convolutional layers followed by ReLU activations.
Then the phase component $\tilde{\mathcal{P}}(X_{t+i})$ of the supporting frames and aggregated amplitude component $\bar{\mathcal{A}}(X)$ are recombined by the inverse Fourier transform $\mathcal{F}^{-1}$. The process is formulated as:
\begin{equation}
    \tilde{x}_{t+i}=\mathcal{F}^{-1}(\bar{\mathcal{A}}(X),\tilde{\mathcal{P}}(X_{t+i})).
\end{equation}
Then the supporting frames are warped to the reference frames using a deformable convolution network (DCN)~\cite{daiDeformableConvolutional2017}. 
The offsets $\delta_{x_{t+i}}, i\in [-N,N], i\ne 0$ are learned from corresponding features $\tilde{x}_{t+i}$ and $\tilde{x}_t$ for the DCN. 
The aligned phase component $\tilde{P}(Y_t)$ is obtained with the learned offsets. 
The process can be formulated as:
\begin{equation}
    \delta_{x_{t+i}}=\mathcal{M}_i([\tilde{x}_{t+i},\tilde{x}_t]),\quad Y_{t+i}=G_i(DCN(\tilde{x}_{t+i},\delta_{x_{t+i}})),
\end{equation}
where $\mathcal{M}_i$ and $\mathcal{G}_i$ are the mapping functions composed of convolution layers. 
This deformable alignment further refines spatial correspondence at the feature level, compensating for residual misalignment after Fourier-domain normalization.
Finally, the reconstructed frames are spatially aligned to the reference frame. The aligned input features ${Y}_{[t-N,t+N]}$ serve as inputs to subsequent reflectance learning, effectively reducing exposure variance in dynamic videos.

\subsection{Exposure Adaptive Fusion}
\label{subsec:exposure_adaptive_fusion}
Although MFA substantially improves feature alignment, small misalignments remain after DCN-based deformable alignment. 
These residual discrepancies can propagate through convolutional layers, leading to overlay artifacts in naive feature fusion. 
The Exposure Adaptive Fusion (EAF) module is designed to mitigate residual misalignments and exposure-induced feature inconsistency through exposure-invariant normalization and inverted attention-guided fusion, adaptively compensating for missing or unreliable details across frames.

To address this issue, we introduce an Exposure Adaptive Fusion (EAF) module, as shown in Fig.~\ref{fig:EAF_module}, which explicitly guides multi-frame fusion using exposure-aware inverted attention.
Specifically, given the aligned features ${Y}_{[t-N:t+N]}$, we first isolate frame-wise statistics by concatenating features along the batch dimension and apply instance normalization~\cite{ulyanovImprovedTexture2017} to extract ``exposure-invariant'' representations.

Different from the Exposure Normalization module in~\cite{huangExposureNormalization2022}, which targets single images, our VECNet++ needs to handle temporal feature interactions in videos. 
Directly applying image-level normalization to stacked video features preserves motion-induced artifacts, as shown in Fig.~\ref{fig:EAF_module} (a). 
To overcome this limitation, we leverage the idea of inverted attention~\cite{huangImprovingObject2020}, which balances feature discrimination and diversity. 
Unlike IAN~\cite{huangImprovingObject2020}, which relies on classification gradients, such signals are unavailable in VEC.
Instead, we calculate the differences $\tilde{F}_{d}$ between the reference frame feature $\tilde{F}_{r}$ and supporting frame features $\tilde{F}_{s}$ as the pseudo gradients, capturing inter-frame feature discrepancies induced by exposure variations to guide selective fusion.
Meanwhile, the common and discriminated features are both important for detail preservation in video enhancement. 
By integrating these complementary cues, the model can robustly enhance both fine regional textures and global exposure consistency across frames.
Finally, these cues are concatenated to generate the inverted attention map $M_{attn}$.
The whole pipeline is:
\begin{equation}
    \label{eq:inverted_attn_map}
    \begin{aligned}
    \mathcal{S}_t &:= \{\tilde{\mathcal{Y}}_{t+i} \mid i\in[-N,N],\, i\neq 0\}, \\
    \tilde{F}_{s} &= \phi(\mathcal{S}_t),\\
    \tilde{F}_{r} &= \phi(\tilde{\mathcal{Y}}_{t}),\\
    \tilde{F}_{d} &= \tilde{F}_{s} - \tilde{F}_{r},\\
    {M}_{\mathrm{attn}} &= 
    \sigma\!\Big(
        \phi\big(
            [\psi(\tilde{F}_r),\,\psi(\tilde{F}_s),\,\psi(\tilde{F}_d)]
        \big)
    \Big),
    \end{aligned}
\end{equation}
where $\phi$ denotes a $1\times1$ convolution, $\psi$ is a pooling operator, and $\sigma(\cdot)$ is the sigmoid function.
After obtaining the inverted attention map $M_{\mathrm{attn}}$, we multiply it with the supporting frame features $\tilde{F}_{s}$. 
The attention-weighted supporting features are then fused with the reference features via an exposure compensation block, as defined in Eq.~\ref{eq:exposure_adaptive_fusion}, effectively suppressing misaligned regions while complementing missing details.
\begin{equation}
\label{eq:exposure_adaptive_fusion}
\tilde{F}_{fusion} = \mathrm{EC}([\tilde{F}_r,\, \tilde{F}_s \odot M_{\mathrm{attn}}]).
\end{equation}

\subsection{Dual-stream Illumination Construction}
\label{subsec:dual_illu_construction}
To explicitly address the asymmetric characteristics of underexposed and overexposed regions, we introduce a dual-stream illumination construction (DIC) unit. By constructing separate illumination representations for underexposure and overexposure, the DIC unit enables disentangled yet complementary modeling of illumination variations.

In Retinex theory, the reference image can be decomposed as $\mathcal{I} = L \odot R$, where $L$ and $R$ denote illumination and reflectance, respectively.
However, standard Retinex-based formulations are primarily effective for enhancing underexposed regions and often struggle to suppress overexposure, where illumination saturation leads to information loss.
Consequently, existing Retinex-based methods~\cite{caiRetinexformerOnestage2023,fuYouNot2023} may tend to exacerbate overexposed regions.

To suppress overexposure while enhancing underexposure, we introduce an antithetic illumination stream by defining an inverted frame $\mathcal{I}_t^{inv} = 1-\mathcal{I}_t$. 
Both $\mathcal{I}_t$ and $\mathcal{I}_t^{inv}$ are processed using shared lightweight pyramid convolutions to estimate latent illumination representations. 
Compared with the original Dual Illumination Construction (DIC) in VECNet~\cite{liuLearningExposure2024}, we simplify the architecture to reduce computational overhead.
Specifically, the original U-Net~\cite{ronnebergerUNetConvolutional2015} is removed, and a compact pyramid convolution structure is adopted to learn the mapping from $\{\mathcal{I}_t, \mathcal{I}_t^{inv}\}$ to $\{\tilde{z}_t, \tilde{z}_t^{inv}\}$ in the latent space.
To ensure reliable illumination generation, we decode the first fine-resolution features $\{z_t, z^{inv}_t\}$ to $\{L_t, L^{inv}_t\}$ as the illumination maps in the image space. 
The enhanced underexposed and overexposed frames are then obtained as:
\begin{equation}
    I_t^u=I_t\odot L_t^{-1}, I_t^{inv}=1-I_t, I_t^o=1-I_t^{inv}\odot(L_t^{inv})^{-1},
\end{equation}
where $I_t^u$ represents the enhanced frame of underexposure, and $\odot$ denotes element-wise multiplication. 
This dual-stream illumination construction enables symmetric handling of under- and overexposure while maintaining low computational cost.

\subsection{Two-stage Synthesis Restoration}
\label{subsec:twostage_synthesis_restoration}
To progressively reconstruct a visually consistent and well-exposed frame, we adopt a two-stage synthesis strategy that decouples illumination-conditioned enhancement from subsequent global refinement.
This design explicitly separates exposure-aware reconstruction from residual artifact correction and consistency refinement, which are non-trivial to jointly address in a single-stage synthesis framework.

The reflectance sub-network $\mathcal{R}$ maps the fused features $\tilde{F}_{fusion}$ to a reflectance representation.
It is implemented using residual channel attention blocks (RCABs)~\cite{zhangImageSuperResolution2018} along with downsampling operations, which facilitate capturing high-level structural information while preserving fine-grained textures through channel-wise reweighting.

In the first synthesis stage, the network $\mathcal{T}_1$ combines the reflectance features with the dual illumination embeddings $\{\tilde{z}_t, \tilde{z}_t^{inv}\}$ to generate an intermediate enhanced result $\tilde{I}_t$.
This stage focuses on exposure-conditioned reconstruction by explicitly injecting illumination priors.
The architecture of $\mathcal{T}_1$ consists of RCABs followed by upsampling layers, with skip connections linking corresponding layers in $\mathcal{R}$ and $\mathcal{T}_1$ to facilitate information reuse and stabilize optimization.

While $\tilde{I}_t$ provides an exposure-aware intermediate result, it may still contain residual artifacts caused by imperfect illumination estimation or motion-induced inconsistencies.
To address this issue, the second synthesis stage $\mathcal{T}_2$ jointly integrates the underexposed enhancement $I_t^u$, the overexposed enhancement $I_t^o$, and the intermediate result $\tilde{I}_t$.
Non-local blocks~\cite{wangNonlocalNeural2018} are incorporated to capture long-range spatial dependencies and improve global consistency.
A learned weight map is then used to adaptively fuse these complementary inputs and regress the final enhanced output $\mathcal{O}_t$.
\begin{equation}
\label{eq:two_stage}
\tilde{I}_t=\mathcal{T}_1([\tilde{z}_t, \tilde{z}_t^{inv}, \mathcal{R}(\tilde{F}_{fusion})]),\quad
\mathcal{O}_t=\mathcal{T}_2(I_t^u,I_t^o,\tilde{I}_t).
\end{equation}

\begin{table*}[!t]
\centering
\caption{Quantitative comparison of under-exposed (Under) and over-exposed (Over) conditions on our proposed DIMEPlus dataset. Best and second-best results are \textbf{bolded} and \underline{underlined}, respectively. Metrics on Under/Over are computed on under-/over-exposed samples. $\uparrow$/$\downarrow$ denotes that higher/lower values are better.}
\label{tab:quantitative_dimeplus}
\begin{tabular}{cc|S[table-format=2.2,mode=text]cc|ccc|ccc}
\toprule
\multirow{2}{*}{Method} &\multirow{2}{*}{Publication} & \multicolumn{3}{c}{Under} & \multicolumn{3}{|c}{Over} & \multicolumn{3}{|c}{Average} \\
\cmidrule(lr){3-5} \cmidrule(lr){6-8} \cmidrule(lr){9-11}
&&  \text{PSNR$\uparrow$} & SSIM$\uparrow$&LPIPS$\downarrow$& PSNR$\uparrow$ & SSIM$\uparrow$&LPIPS$\downarrow$& PSNR$\uparrow$ & SSIM$\uparrow$&LPIPS$\downarrow$\\
\midrule
RVRT~\cite{liangRecurrentVideo2022} &NeurIPS 2022&15.23&0.7831&0.3243&17.26&0.8592&0.1868&16.22&0.8201&0.2575\\
SI-HDR~\cite{leSingleImageHDR2023} & WACV 2023 &13.61&0.7363&0.4541&16.86&0.8309&0.2591&15.19&0.7822&0.3594\\
\midrule
SDSDNet~\cite{wangSeeingDynamic2021} & ICCV 2021&14.91&0.6853&0.3608&18.89&0.8820&0.2031&16.84&0.7808&0.2842\\
Retinexformer~\cite{caiRetinexformerOnestage2023} & ICCV 2023&17.47&0.7836&0.2651&20.08&0.8783&0.1823&18.74&0.8296&0.2249\\
GTMeanLoss~\cite{liaoGTMeanLoss2025} (with \cite{caiRetinexformerOnestage2023})& ICCV 2025&16.59&0.8108&\underline{0.2141}&19.57&0.8953&\underline{0.1401}&18.04&0.8518&\underline{0.1782}\\
HVI-CIDNet~\cite{yanHVINew2025} & CVPR 2025&17.93&0.7660&0.2639&21.99&0.9140&0.1622&19.90&0.8378&0.2145\\
\midrule
IAT~\cite{cuiYouOnly2022}&BMVC2022&12.89&0.6421&0.4907&19.75&0.8853&0.1920&16.22&0.7602&0.3457\\
DRBN-ENC~\cite{huangExposureNormalization2022} & CVPR 2022&17.38&0.7836&0.2355&20.66&0.9112&0.1567&18.97&0.8455&0.1973\\
ECLNet~\cite{huangExposureConsistencyRepresentation2022} &ACM MM 2022&16.60&0.7794&0.2496&20.48&0.8569&0.2093&18.48&0.8170&0.2301\\
FECNet~\cite{huangDeepFourierBased2022} &ECCV 2022&17.52&0.7618&0.2857&22.13&0.9132&0.1642&19.76&0.8353&0.2267\\
LCDP~\cite{wangLocalColor2022}&ECCV 2022 &\underline{19.13}&0.7564&0.2648&21.62&0.8823&0.1740&20.34&0.8175&0.2207\\
LACT~\cite{baekLuminanceawareColor2023b}&ICCV 2023&17.94&0.7714&0.2457&22.69&0.8983&0.1583&20.25&0.8330&0.2033\\
MSLT++~\cite{zhou4KResolutionPhoto2024} &  WACV 2024&14.19&0.6100&0.5410&21.11&0.8857&0.1820&17.55&0.7438&0.3667\\
RECNet~\cite{liuRegionAwareExposure2024} &  AAAI 2024&17.47&\underline{0.8267}&0.2577&19.33&0.8889&0.2078&18.37&0.8569&0.2335\\
CSEC~\cite{liColorShift2024} &  CVPR 2024&17.20&0.7515&0.2973&20.27&0.8659&0.1849&18.69&0.8070&0.2428\\
CoTF~\cite{liRealTimeExposure2024} &  CVPR 2024&17.91&0.7751&0.2775&21.64&0.9011&0.1700&19.72&0.8363&0.2253\\
ExposureSlot~\cite{jungExposureslotExposurecentric2025a} &  CVPR 2025&16.39&0.7672&0.2828&18.41&0.8651&0.1998&17.37&0.8147&0.2425\\
\midrule
VECNet~\cite{liuLearningExposure2024} & ACM MM 2024& 18.25 &0.8168&0.2688&\underline{22.76}&\underline{0.9188}&0.1814&\underline{20.44}&\underline{0.8663}&0.2264\\
\textbf{VECNet++} & - & \textbf{20.86}&\textbf{0.8672}&\textbf{0.1943}&\textbf{22.88}&\textbf{0.9197}&\textbf{0.1390}&\textbf{21.84}&\textbf{0.8927}&\textbf{0.1674}\\
\bottomrule
\end{tabular}
\end{table*}

\begin{figure*}[!t]
    \centering
    \includegraphics[width=1.0\linewidth]{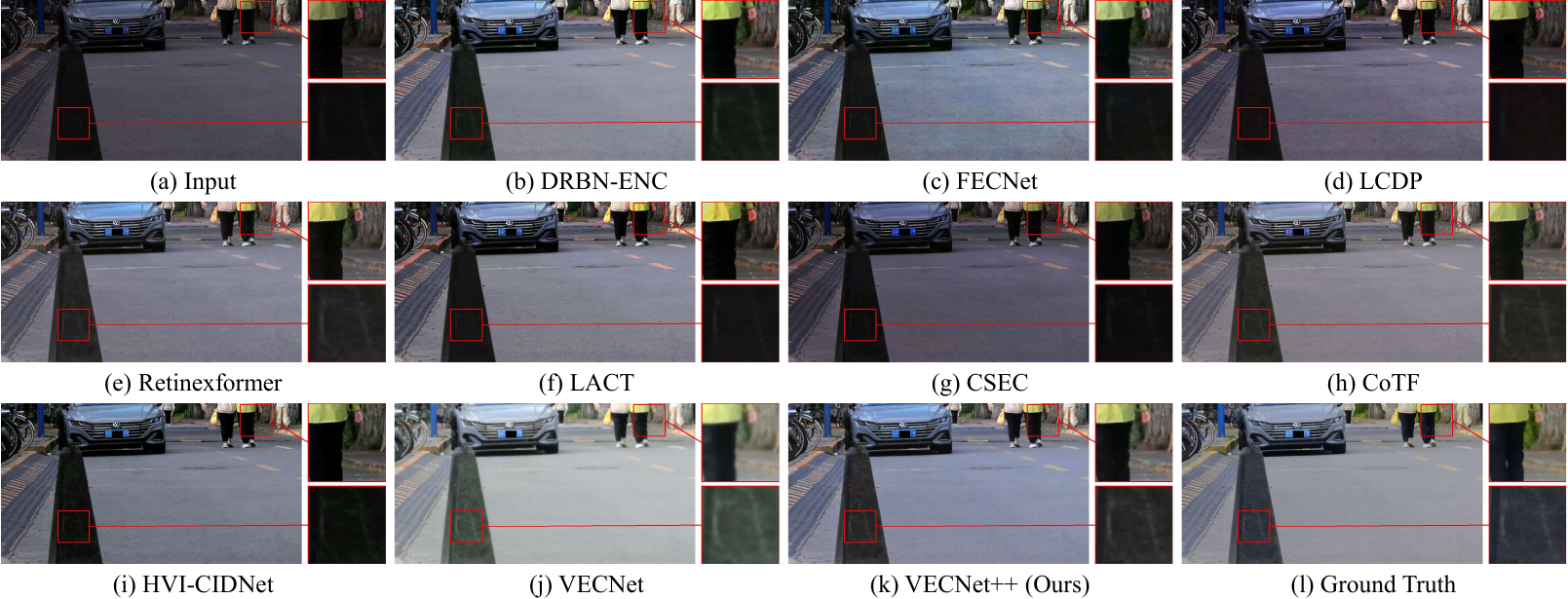}
    \caption{Visual comparisons with different methods on the DIMEPlus dataset. Our method (k) achieves better detail reconstruction.}
    \label{fig:comparison_DIMEPlus}
\end{figure*}

\subsection{Objectives}
We incorporate a range of loss terms to facilitate the training of our model. We employ the commonly used pixel-wise charbonnier loss term, denoted as $\mathcal{L}_{pix}$, to measure the accuracy of pixel reconstruction. To guide the network to estimate and smooth the dual-stream illumination, we apply the total variation~\cite{wangUnderexposedPhoto2019} loss term $\mathcal{L}_{tv}$:
\begin{equation}    
    \mathcal{L}_{tv}=\sum_c[\left(\partial_{x/y}L_t\right)^2+\left(\partial_{x/y}L_t^{inv}\right)^2],
\end{equation}
where all channels ($c$) of all pixels are summed, $\partial_{x/y}$ are partial derivatives in horizontal and vertical directions in the image space. 
To encourage exposure consistency in the reconstructed frame, we introduce an amplitude consistency loss term $\mathcal{L}_{amp}$, which can be formulated as:
\begin{equation}
    \mathcal{L}_{amp}=||\mathcal{A}(X_{O_t})-\mathcal{A}(X_{I_t^{gt}})||_F^1,
\end{equation}
where $X_{(\cdot)}$ represents the frequency component of output $\mathcal{O}_t$ or ground truth normal-exposed image $I_t^{gt}$ in Fourier space.
In addition, to recover the lost color details by the instance normalization and enforce color consistency between the enhanced result and the ground truth, we compute a chromaticity-aligned loss $\mathcal{L}_{color}$ in the XYZ color space, which can be formulated as Eq.~\ref{eq:color_loss}.
This loss penalizes hue shifts and effectively stabilizes color reproduction under varying exposure conditions.
\begin{equation}
    \label{eq:color_loss}
    \begin{aligned}
    \mathcal{L}_{\mathrm{color}}=\frac1N\sum_{i=1}^N\left\|\hat{\mathcal{O}}_{t} - \hat{\mathcal{I}}^{gt} \right\|_2^2, \quad \hat{\mathcal{I}}=\mathbf{T}\cdot\mathcal{I},
    \end{aligned}
\end{equation}
where $\mathbf{T}$ denotes the linear transformation matrix from RGB to XYZ color space and $N$ is the number of pixels.

The overall function can be written as:
\begin{equation}
    \label{eq:whole_loss}
    \mathcal{L}_{total}=\lambda_1\mathcal{L}_{pix}+\lambda_2\mathcal{L}_{tv}+\lambda_3\mathcal{L}_{amp}+\lambda_4\mathcal{L}_{color},
\end{equation}
where $\lambda_1$, $\lambda_2$, $\lambda_3$, and $\lambda_4$ are weight hyperparameters.

\section{Experiments}
\label{sec:exp}
In this section, we conduct comprehensive experiments on the proposed DIMEPlus dataset, along with various other real-world datasets. 
The results validate the effectiveness of the dataset and the superiority of our proposed method for video exposure correction.

\begin{figure*}[!t]
    \centering
    \includegraphics[width=1.0\linewidth]{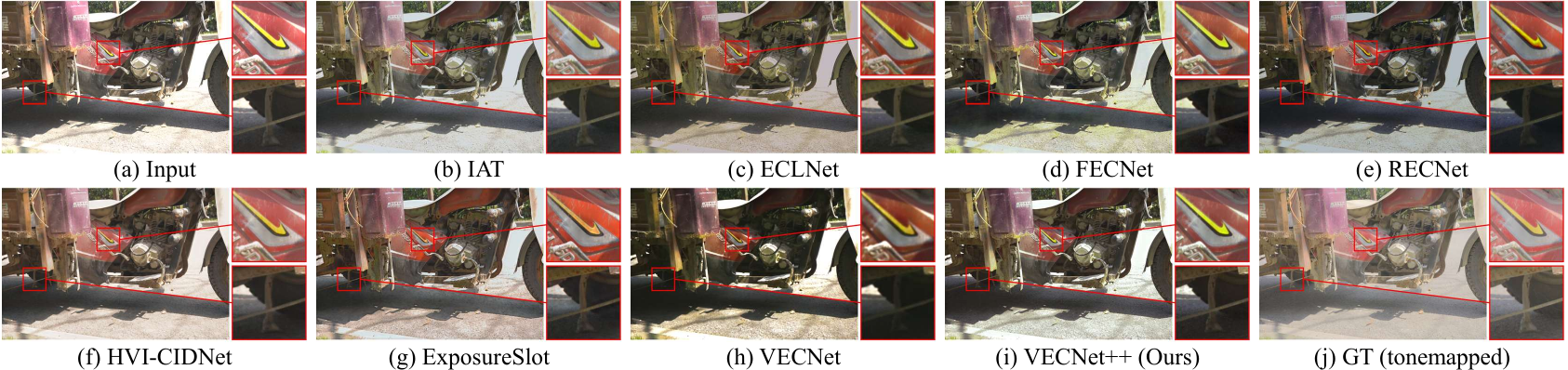}
    \caption{Qualitative comparisons with different methods on the RealHDRV dataset.}
    \label{fig:comparison_RealHDRV}
    \vspace{-1em}
\end{figure*}

\begin{table}[!t]
\centering
\caption{Quantitative results of different methods on Real-HDRV~\cite{shuRealworldHDR2024} dataset.
Best and second-best results are \textbf{bolded} and \underline{underlined}, respectively. $\uparrow$/$\downarrow$ denotes that higher/lower values are better.}
\label{tab:02_quantitative_realhdrv}
\begin{tabular}{l|ccc}
\toprule
Method&PSNR$\uparrow$ & SSIM$\uparrow$ & LPIPS$\downarrow$ \\
\midrule
RVRT~\cite{liangRecurrentVideo2022} &15.60&0.7642&0.1917\\
SI-HDR~\cite{leSingleImageHDR2023} &13.99&0.7177&0.3262\\
\midrule
SDSDNet~\cite{wangSeeingDynamic2021}&12.44&0.5716&0.3729\\
Retinexformer~\cite{caiRetinexformerOnestage2023}&16.63&0.7793&0.1863\\
GTMeanLoss~\cite{liaoGTMeanLoss2025} (with \cite{caiRetinexformerOnestage2023})&16.19&0.7606&0.2018\\
HVI-CIDNet~\cite{yanHVINew2025} &17.37&\underline{0.8178} &0.2099\\
\midrule
IAT~\cite{cuiYouOnly2022}& 17.63&0.7741 & 0.1951 \\
DRBN-ENC~\cite{huangExposureNormalization2022}&16.08&0.7626 & 0.2147 \\
ECLNet~\cite{huangExposureConsistencyRepresentation2022}&17.55&0.7821 & 0.2350 \\
FECNet~\cite{huangDeepFourierBased2022} &16.89&0.7747 & 0.2170 \\
LCDP~\cite{wangLocalColor2022}&16.24&0.7416 & 0.2008 \\
LACT~\cite{baekLuminanceawareColor2023b}&15.33&0.7279 & 0.2007 \\
MSLT++~\cite{zhou4KResolutionPhoto2024} &16.59&0.7791 & 0.1886 \\
RECNet~\cite{liuRegionAwareExposure2024}&17.77&0.8170 & 0.2034 \\
CSEC~\cite{liColorShift2024} &15.04&0.7113 & 0.2161 \\
CoTF~\cite{liRealTimeExposure2024}&15.86&0.7660& 0.2012 \\
ExposureSlot~\cite{jungExposureslotExposurecentric2025a}&\underline{18.07}&0.8125& \underline{0.1804} \\
\midrule
VECNet~\cite{liuLearningExposure2024} &17.61&0.7993&0.2009\\
\textbf{VECNet++} &\textbf{18.85}&\textbf{0.8298}&\textbf{0.1775}\\
\bottomrule
\end{tabular}
\end{table}

\subsection{Datasets} 
\textbf{Our Proposed DIMEPlus}. 
To evaluate the proposed method, we conduct quantitative experiments on the testing split of DIMEPlus, which contains 15 underexposed and 15 overexposed video pairs.
All compared baseline methods are trained solely on DIMEPlus without any additional fine-tuning or retraining on other datasets, ensuring a fair comparison.

\textbf{Real-HDRV}~\cite{shuRealworldHDR2024} is originally designed for HDR video reconstruction rather than sRGB-domain video exposure correction. 
To obtain reference videos for evaluation, we apply the widely used \textit{tonemapReinhard} operator to convert HDR videos into tone-mapped sRGB counterparts.
The Real-HDRV test set consists of 50 video sequences with paired under- and overexposed inputs, totaling 800 frames.

\textbf{LLIV-Phone}~\cite{liLowLightImage2022} is a large-scale unpaired real-world low-light video dataset, containing 120 videos with 45,148 frames captured by diverse mobile devices.
Due to the large scale of the dataset and the fact that it only includes low-light samples, we follow the evaluation protocol of SDSDNet and use the first 30 frames of each video for testing, totaling 3,600 frames.

\textbf{ProCaptureVideo Dataset.}
We further collect a real-world video dataset for cross-dataset generalization evaluation, primarily sourced from Pexels\footnote[1]{Pexels. \url{https://www.pexels.com/}}.
This dataset contains 15 videos captured by professional videographers, with frame rates ranging from 24 to 60 fps and resolutions spanning from $1280\times720$ to $2048\times1080$, covering diverse real-world scenes.

\subsection{Implementation Details}
We implement our framework with PyTorch on a single NVIDIA GeForce RTX 3090 GPU. 
We use $N=2$ consecutive frames without intervals as input. 
The DIMEPlus dataset is split into training, validation, and testing sets following a ratio of 153:9:30 videos. The model is trained for a total of 800,000 iterations.
We adopt the Adam optimizer with $\beta_1=0.9$ and $\beta_2=0.99$.
The learning rate is set to $1\times10^{-3}$, and the batch size is 16.
Training is performed on cropped patches of size $256\times256$.
The weights for the terms in the loss function in Eq.~\ref{eq:whole_loss} are set as $\lambda_1,\lambda_2,\lambda_3,\lambda_4=1.0, 0.01, 100.0, 0.01$.

\begin{table*}[!t]
\centering
\begin{threeparttable}
\caption{Quantitative results of different methods on the DIMEPlus, LLIV-Phone~\cite{liLowLightImage2022}, and ProCaptureVideo datasets with no-reference metrics. $\downarrow$ denotes that lower values are better.}
\label{tab:dimeplus_lliv_phone_noref}
\begin{tabular}{l|
    S[table-format=1.2,mode=text]
    S[table-format=2.2,mode=text]
    S[table-format=1.4,mode=text]
    S[table-format=3.2,mode=text]|
    S[table-format=1.2,mode=text]
    S[table-format=2.2,mode=text]
    S[table-format=1.4,mode=text]
    r|
    S[table-format=1.2,mode=text]
    S[table-format=2.2,mode=text]
    S[table-format=1.4,mode=text]
    S[table-format=2.2,mode=text]}
\toprule
\multirow{2}{*}{Method} & \multicolumn{4}{c|}{\text{DIMEPlus}} & \multicolumn{4}{c|}{\text{LLIV-Phone~\cite{liLowLightImage2022}}}& \multicolumn{4}{c}{\text{ProCaptureVideo}} \\ \cmidrule(lr){2-5} \cmidrule(lr){6-9} \cmidrule(lr){10-13}
& \text{NIQE}$\downarrow$ & \text{ILNIQE}$\downarrow$ & \text{BE}$\downarrow$ & \text{ALV}$\downarrow$  & \text{NIQE}$\downarrow$ & \text{ILNIQE}$\downarrow$ & \text{BE}$\downarrow$ & \text{ALV}$\downarrow$ & \text{NIQE}$\downarrow$ & \text{ILNIQE}$\downarrow$ & \text{BE}$\downarrow$ & \text{ALV}$\downarrow$ \\
\midrule
RVRT~\cite{liangRecurrentVideo2022} &5.09&34.11&0.2331&37.86&10.53&44.22&0.3202&5.20 &6.37&36.66&0.2705&2.70\\
SI-HDR~\cite{leSingleImageHDR2023} &5.99&43.18&\underline{0.0867}&132.79&8.57&60.09&0.2568&187.28&5.68&39.49&0.1262&94.55\\
\midrule
SDSDNet~\cite{wangSeeingDynamic2021} &{\underline{\num{4.70}}}&33.07&0.1442&65.92&\underline{6.04}&45.68&0.3586&7.69&4.54&30.40&0.2435&2.51\\
Retinexformer~\cite{caiRetinexformerOnestage2023}&5.08&\underline{30.73}&0.0976&27.97&6.11&33.94&0.2354&3.56&4.08&25.61&0.1429&11.06\\
GTMeanLoss~\cite{liaoGTMeanLoss2025}&5.08&32.44&0.1290&38.65&6.13&36.95&0.2916&\underline{3.53}&4.23&25.58&0.1447&3.39\\
HVI-CIDNet~\cite{yanHVINew2025} &4.99&32.34&0.1084&42.04&7.34&43.83&0.3006&3.75&4.31&26.80&0.1589&2.88\\
\midrule
IAT~\cite{cuiYouOnly2022}&4.71&31.71&0.1083&71.81&6.89&36.08&0.2734&8.76&4.21&25.76&0.1580&5.73\\
DRBN-ENC~\cite{huangExposureNormalization2022} &5.81&37.33&0.1242&41.83&7.56&42.34&0.2945&10.71&5.06&27.47&0.1528&4.44\\
ECLNet~\cite{huangExposureConsistencyRepresentation2022} &5.45&32.59&0.1307&65.63&6.21&\underline{33.06}&0.2391&11.28&4.90&25.22&0.1402&7.11\\
FECNet~\cite{huangDeepFourierBased2022} &5.20&32.52&0.1269&37.62&6.65&35.69&0.2581&6.63&4.45&26.30&0.1536&5.12\\
LCDP~\cite{wangLocalColor2022}&4.85&32.78&0.1151&50.25&7.91&41.88&0.3069&18.72&4.15&27.05&0.1688&13.29\\
LACT~\cite{baekLuminanceawareColor2023b}&4.72&30.82&0.1380&76.08&6.11&34.94&0.2674&13.90&4.31&\underline{25.00}&0.1430&6.13\\
MSLT++~\cite{zhou4KResolutionPhoto2024} &4.85&33.35&0.1354&45.10&7.79&42.27&0.2929&15.09&4.13&29.00&\underline{0.1055}&56.48\\
RECNet~\cite{liuRegionAwareExposure2024}&5.15&32.27&0.0911&79.89&5.94&36.62&0.2871&46.77&4.70&26.35&0.1521&10.28\\
CSEC~\cite{liColorShift2024}&4.75&31.22&0.1511&95.89&7.95&38.46&0.3230&3.56&4.25&27.86&0.2038&16.76\\
CoTF~\cite{liRealTimeExposure2024} &4.87&31.72&0.1166&43.15&8.79&43.07&0.3070&8.25&\underline{3.86}&26.30&0.1568&10.35\\
ExposureSlot~\cite{jungExposureslotExposurecentric2025a}&4.75&31.49&0.1174&51.58&6.10&38.76&0.2786&12.32&3.95&25.15&0.1276&5.69\\
\midrule
VECNet~\cite{liuLearningExposure2024} &4.86&32.46&0.1355&\underline{27.18} &6.92&39.82&\underline{0.2310}&4.95&4.36&25.04&0.1354&\underline{1.96}\\
\textbf{VECNet++}&\textbf{4.63}&\textbf{30.67}&\textbf{0.0695}&\textbf{20.95}&\textbf{5.81}&\textbf{33.04}&\textbf{0.2238}&\textbf{3.43}&\textbf{3.81}&\textbf{24.00}&\textbf{0.1028}&\textbf{1.48}\\
\bottomrule
\end{tabular}
\begin{tablenotes}
\footnotesize
\item * GTMeanLoss~\cite{liaoGTMeanLoss2025} is implemented based on \cite{caiRetinexformerOnestage2023}.
\end{tablenotes}
\end{threeparttable}
\vspace{-1em}
\end{table*}

\begin{figure*}[!t]
    \centering
    \includegraphics[width=1.0\linewidth]{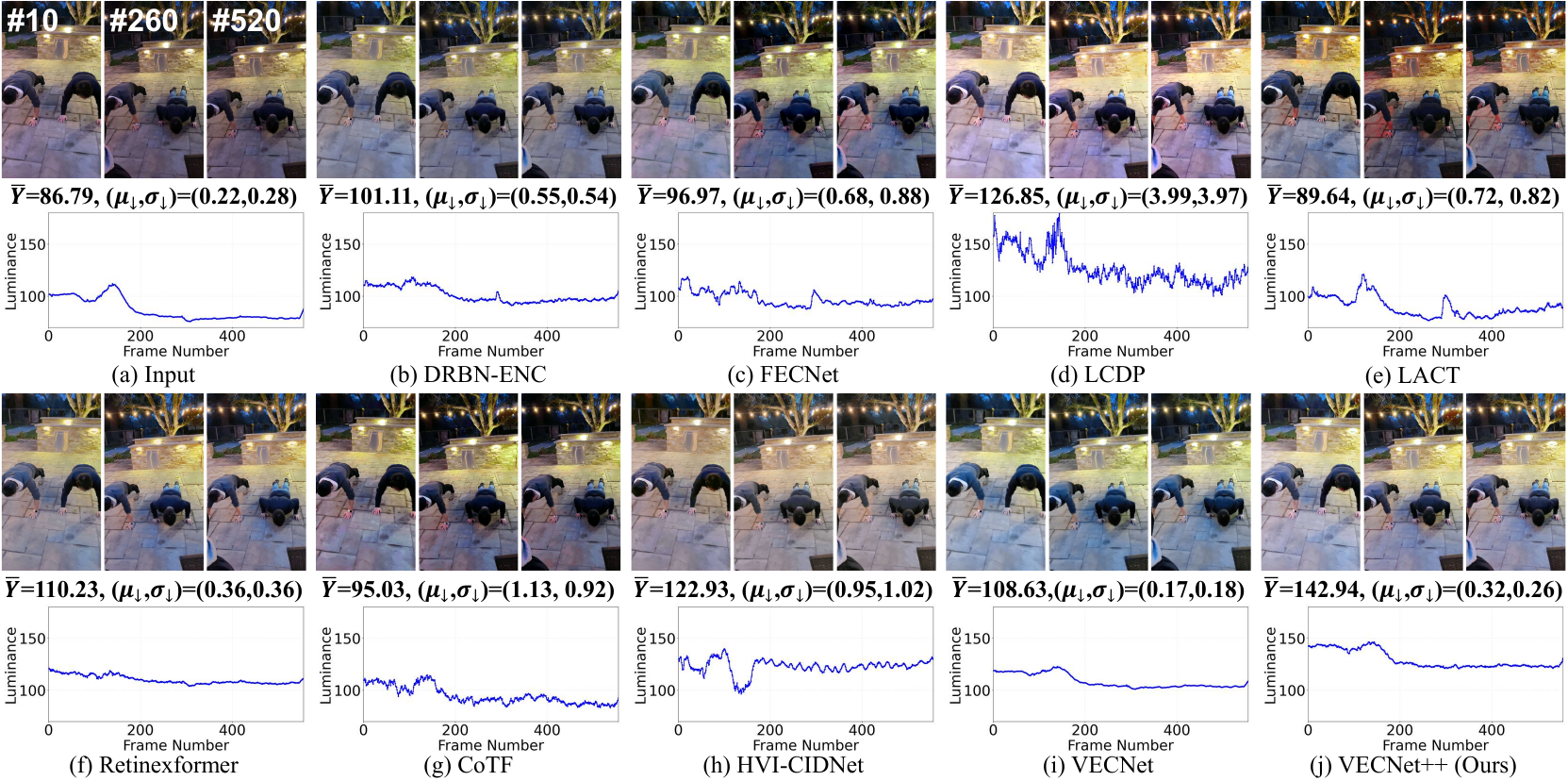}
    \caption{Qualitative comparisons with different methods on the LLIV-Phone dataset. For each method, the top row shows the corrected results, while the bottom row presents the corresponding luminance curves over the entire video. $\bar{Y}$ denotes the mean luminance, reflecting the overall strength of exposure correction. $(\mu_\downarrow, \sigma_\downarrow)$ represent the mean and standard deviation of inter-frame differences, respectively, where $\downarrow$ indicates lower is better.}
    \label{fig:comparison_LLIVPhone}
    \vspace{-1em}
\end{figure*}

\subsection{Baselines}
Since few prior works explicitly address video exposure correction, we compare our method with a comprehensive set of state-of-the-art image-based exposure correction approaches, including
IAT~\cite{cuiYouOnly2022},
DRBN-ENC~\cite{huangExposureNormalization2022},
ECLNet~\cite{huangExposureConsistencyRepresentation2022},
FECNet~\cite{huangDeepFourierBased2022},
LCDP~\cite{wangLocalColor2022},
LACT~\cite{baekLuminanceawareColor2023b},
MSLT++~\cite{zhou4KResolutionPhoto2024},
RECNet~\cite{liuRegionAwareExposure2024},
CSEC~\cite{liColorShift2024},
CoTF~\cite{liRealTimeExposure2024},
and ExposureSlot~\cite{jungExposureslotExposurecentric2025a}.
These methods focus on exposure correction but operate exclusively at the image level without temporal modeling.

To further enrich the evaluation, we additionally include representative methods from related low-level vision tasks such as low-light enhancement and video restoration.
Specifically, we consider SDSDNet~\cite{wangSeeingDynamic2021},
RVRT~\cite{liangRecurrentVideo2022},
Retinexformer~\cite{caiRetinexformerOnestage2023},
SI-HDR~\cite{leSingleImageHDR2023},
HVI-CIDNet~\cite{yanHVINew2025},
and GTMeanLoss~\cite{liaoGTMeanLoss2025}.
Although these methods are not designed for exposure correction, they provide meaningful references for illumination enhancement and temporal restoration.

\subsection{Evaluation Metrics}
We evaluate reconstruction accuracy and perceptual quality using both reference-based and no-reference metrics.
The reference-based metrics include
(i) Peak Signal-to-Noise Ratio (PSNR),
(ii) Structural Similarity Index Measure (SSIM),
and (iii) Learned Perceptual Image Patch Similarity (LPIPS)~\cite{zhangUnreasonableEffectiveness2018a}.
The no-reference metrics include 
(iv) Natural Image Quality Evaluator (NIQE)~\cite{mittalMakingCompletely2013a},
(v) Integrated Local Natural Image Quality Evaluator (ILNIQE)~\cite{zhangFeatureEnrichedCompletely2015},
(vi) Brightness Error (BE)~\cite{chenMinimumMean2003},
and (vii) Average Luminance Variance (ALV)~\cite{liLowLightImage2022}.
These metrics jointly assess accuracy, perceptual quality, and exposure characteristics.

\subsection{Comparison and Evaluation}
\noindent\textbf{Evaluation on the proposed DIMEPlus.} We conduct experiments on the proposed DIMEPlus dataset, and the quantitative results are shown in Tab.~\ref{tab:quantitative_dimeplus}. Our VECNet++ consistently outperforms all image- and video-based enhancement or restoration methods across all metrics, demonstrating its superior capability in video exposure correction. Notably, VECNet++ surpasses the previous VECNet by 1.40 dB in PSNR. Compared with other baseline methods, VECNet++ achieves a substantial improvement of 1.50 dB in PSNR, highlighting its significant performance advantage.
Fig.~\ref{fig:comparison_DIMEPlus} shows the visual comparisons on the DIMEPlus dataset. 
It can be observed that several existing methods fail to recover the texture on the plastic traffic barrier in the foreground, as this region is extremely dark.
The latest exposure correction method, CoTF~\cite{liRealTimeExposure2024}, recovers the detail but exhibits significant hue shifts, especially in the license plate (turning the blue into purple).
In contrast, our VECNet++ alleviates motion-induced blurring artifacts compared with the prior VECNet, and achieves a balanced performance in details preservation and color enhancement.

\begin{figure*}[!t]
    \centering
    \includegraphics[width=1.0\linewidth]{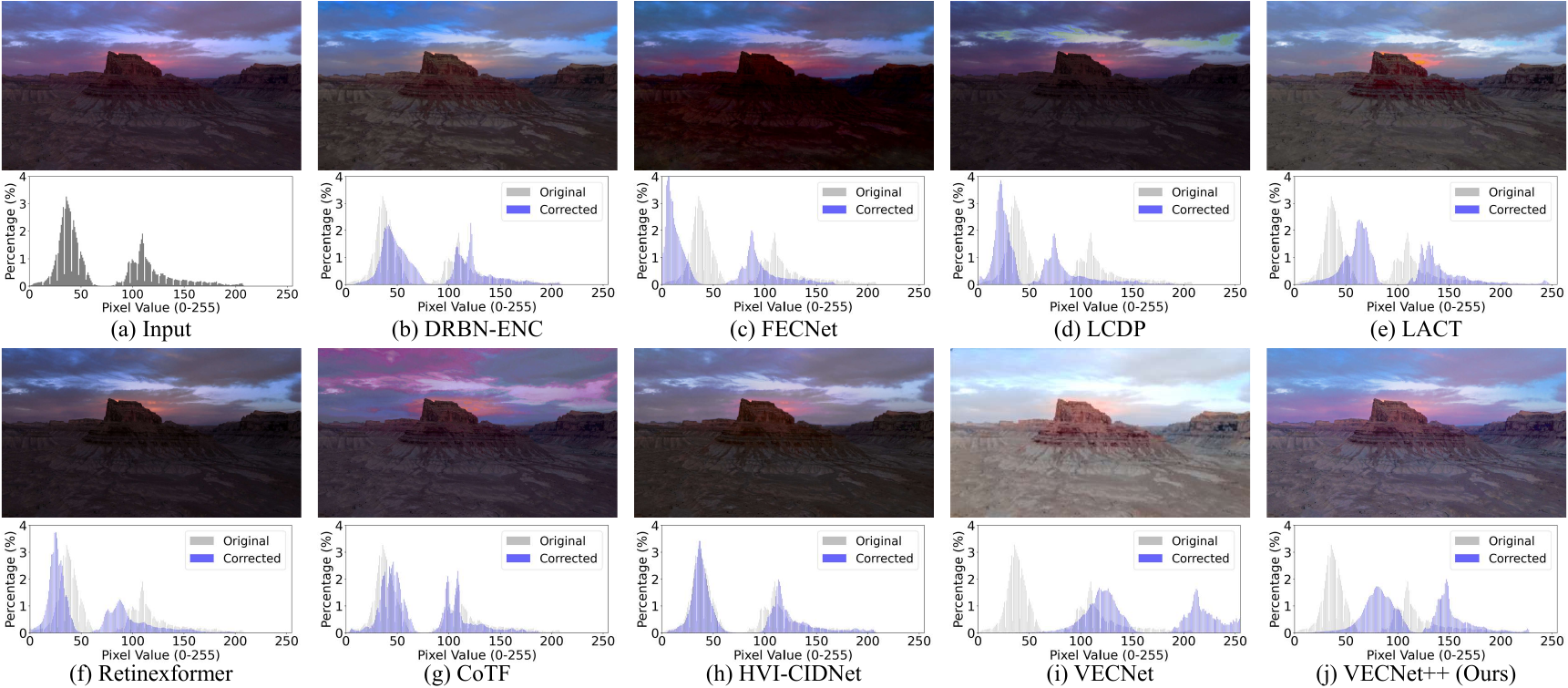} 
    \caption{Qualitative comparisons of different methods on the ProCaptureVideo dataset. For each method, the top row shows the enhanced frames, while the bottom row presents the corresponding intensity histograms. Our method achieves smoother exposure correction with reduced sharp peaks.}
    \label{fig:comparison_Pexels}
    \vspace{-1em}
\end{figure*}

\begin{figure}[!t]
    \centering
    \includegraphics[width=1.0\linewidth]{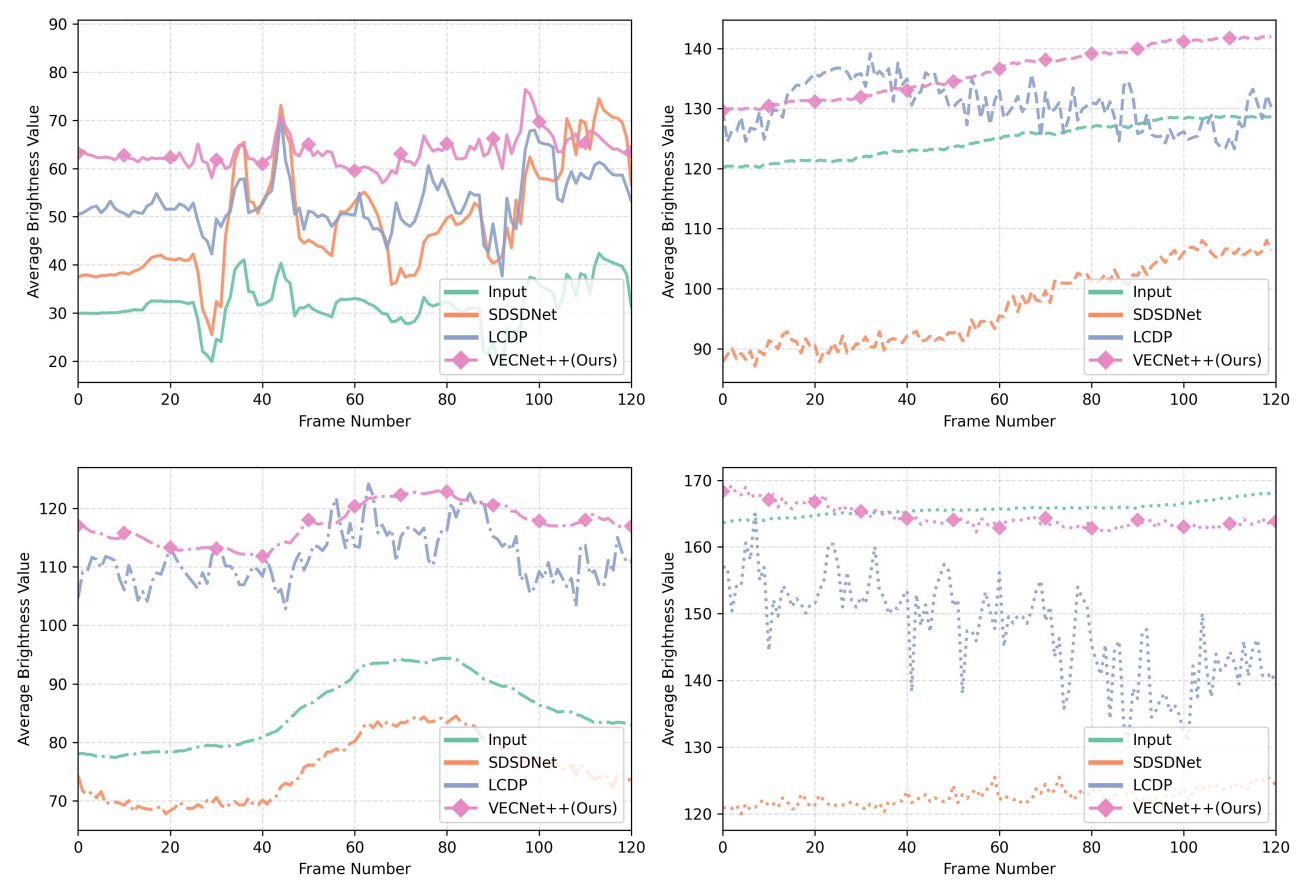}
    \caption{Video brightness curves of different enhancement methods. Each subplot corresponds to a single video. Compared with prior image-based methods, VECNet++ yields smoother curves, indicating improved performance on exposure correction and temporal consistency.}
    \vspace{-1em}
    \label{fig:comparison_brightness_curves}
\end{figure}

\noindent\textbf{Evaluation on challenging diverse exposure conditions.}
To evaluate the generalization ability of our method under challenging dynamic scenes, we conduct extensive experiments on the paired real-world datasets, Real-HDRV~\cite{shuRealworldHDR2024}. Tab.~\ref{tab:02_quantitative_realhdrv} shows the quantitative results. Our VECNet++ surpasses the latest competing method, ExposureSlot~\cite{jungExposureslotExposurecentric2025a} in PSNR of 0.78 dB on Real-HDRV.
Fig.~\ref{fig:comparison_RealHDRV} exhibits the visual comparisons on the Real-HDRV dataset. Our method can achieve a better balance in both underexposed and overexposed regions. It should be noted that Real-HDRV does not explicitly provide the sRGB ground truth, and we use the prevailing tonemap operator \textit{tonemapReinhard} to obtain Fig.~\ref{fig:comparison_RealHDRV} (j), which may introduce additional bias.

\noindent\textbf{Evaluation on temporal consistency reconstruction.} 
To evaluate temporal consistency in real-world video reconstruction, we conduct the experiments on our DIMEPlus and unpaired datasets LLIV-Phone~\cite{liLowLightImage2022} and ProCaptureVideo in metrics of NIQE, ILNIQE, BE, and ALV, as reported in Tab.~\ref{tab:dimeplus_lliv_phone_noref}.
Our method achieves the best results in all metrics, demonstrating stability when handling real-world videos with various motions under challenging exposure conditions. 
Compared with our previous VECNet, the VECNet++ exhibits robust temporal exposure consistency, with the improvements of ALV by 6.23, 1.52, and 0.48 in DIMEPlus, LLIV-Phone, and ProCaptureVideo datasets, respectively.
Compared to other baseline methods, our approach achieves better perceptual quality with consistently lower ALV across all three datasets.

Fig.~\ref{fig:comparison_LLIVPhone} presents qualitative comparisons on the LLIV-Phone dataset, where three frames with a long temporal span are selected for visualization.
In addition to visual results, we plot the luminance curves over time to analyze temporal exposure stability.
To quantify temporal consistency, we compute the inter-frame luminance difference as $\Delta Y_t = \left| \bar{Y}_{t} - \bar{Y}_{t-1} \right|$, where $\bar{Y}_t$ denotes the mean luminance of frame $t$. 
The temporal stability is measured by $\mu$ and $\sigma$, denoting the mean and standard deviation of $\Delta Y_t$ over the entire video, respectively, where lower values indicate better temporal consistency.
Image-based methods do not exploit temporal information, leading to noticeable exposure fluctuations and flickering artifacts.
In contrast, VECNet++ produces visually consistent results with smoother luminance transitions.

Fig.~\ref{fig:comparison_Pexels} shows the qualitative results on the ProCaptureVideo dataset. Our VECNet++ could lighten the underexposed regions on the rocky surface while suppressing the overexposure in the sky. 
In addition, we visualize the brightness curves of selected videos from the ProCaptureVideo dataset, as shown in Fig.~\ref{fig:comparison_brightness_curves}. 
The results demonstrate that our VECNet++ remains robust under drastic brightness variations.

\subsection{Ablation Study and Analysis}
\begin{figure}[t]
    \centering
    \includegraphics[width=1.0\linewidth]{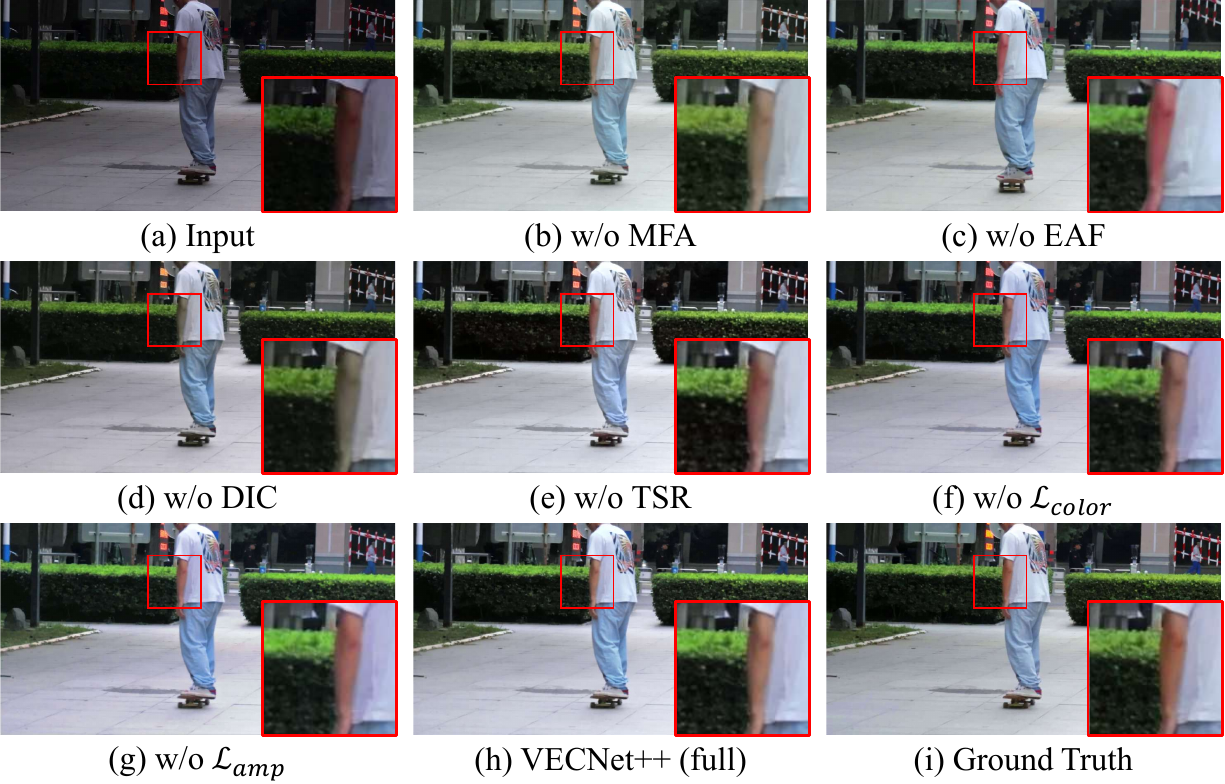}
    \caption{Visual ablation study on the proposed modules.}
    \label{fig:comparison_ablations}
    \vspace{-1em}
\end{figure}

We provide a series of ablation studies on the DIMEPlus dataset to evaluate the effectiveness of each component of the proposed method.
The ablations can be divided into three parts: (a) ablations on core architecture designs, (b) ablations on loss functions, and (c) ablations on auxiliary components. 
The results are shown in Tab.~\ref{tab:ablations} and Fig.~\ref{fig:comparison_ablations}.

\begin{table}[!t]
\centering
\begin{threeparttable}
\caption{Ablation study on architectural designs.}
\label{tab:ablations}
\setlength{\tabcolsep}{2pt}
\begin{tabular}{l|S[table-format=2.3,mode=text]cc}
\toprule
\multirow{2}{*}{Model} &\text{Trainable} &\multirow{2}{*}{PSNR} &\multirow{2}{*}{SSIM} \\
& \text{Params (M)} &&\\
\midrule
\textbf{Full VECNet++ (Ours)}&\textbf{6.786}&\textbf{21.84}&\textbf{0.8927}\\
\midrule
\multicolumn{4}{c}{\textit{Core Architecture Designs}}\\
w/o Multi-frame Fourier Alignment (MFA) &6.780 &21.58&0.8829\\
w/o Exposure Adaptive Fusion (EAF) &6.760&19.57&0.8485\\
w/o MFA and EAF & 6.637 & 19.11 &0.8710\\
w/o Dual-stream Illumination Construction (DIC)&6.189& 18.85&0.8808 \\ 
w/o Two-stage Synthesis Restoration (TSR)&6.438& 20.02&0.8401\\
w/o DIC and TSR & 5.847 &18.44 & 0.8491\\
\midrule
\multicolumn{4}{c}{\textit{Loss Function Ablation}}\\
w/o $\mathcal{L}_{color}$&6.786&20.96&0.8807\\
w/o $\mathcal{L}_{amp}$&6.786&19.93&0.8735\\
\midrule
\multicolumn{4}{c}{\textit{Auxiliary Architectural Improvements}}\\
w/o Optical Flow Matching&6.786& 21.50&0.8446 \\
w/o Efficient Illumination Estimator &10.843&20.05&0.8494 \\  
\midrule
w/o All\textsuperscript{*} &10.102&21.38&0.8945\\
\bottomrule
\end{tabular}
\begin{tablenotes}
    \item w/o All\textsuperscript{*}: All core components (MFA, EAF, DIC, and TSR) together with the flow-based pre-alignment module are removed. The model is trained using only $\mathcal{L}_{pix}$ and employs the original heavy illumination estimator.
\end{tablenotes}
\end{threeparttable}
\end{table}

\textbf{Ablations on Core Architecture Designs.} We conduct ablation studies on the core architectural components by individually removing each module.
For ``w/o MFA'', we bypass Fourier-based alignment and directly feed the preceding multi-frame features into EAF, while keeping the subsequent fusion and restoration pipeline unchanged.
For ``w/o EAF'', the EAF module is excluded to assess the contribution of inverted attention maps in multi-frame exposure-adaptive fusion.
For ``w/o DIC'', only the illumination map from input frame $I_t$ is retained, discarding the illumination map from the inverse input frame $I^{inv}_t$. 
For ``w/o TSR'', the second-stage synthesis module is removed to evaluate the effect of the two-stage reconstruction strategy.
The joint ablation results show that both ``w/o MFA and EAF'' and ``w/o DIC and TSR'' lead to larger PSNR drops than the corresponding single-module ablations, indicating their complementary contributions.
Each module brings expected quantitative performance improvements as reported in Tab.~\ref{tab:ablations}. 
Removing the MFA module (Fig.~\ref{fig:comparison_ablations} (b)) leads to noise and significant motion blurs.
For the EAF module, there exist minor overlay artifacts after removing it, as shown in Fig.~\ref{fig:comparison_ablations} (c).
Removing the DIC module leads to marginal performance drops, indicating its complementary role, from Fig.~\ref{fig:comparison_ablations} (d). 
Comparing Fig.~\ref{fig:comparison_ablations} (e) and the final result (h), our two-stage strategy helps preserve more color and texture information.

\textbf{Ablations on Loss Functions.} We propose the $\mathcal{L}_{amp}$ and $\mathcal{L}_{color}$ to ensure exposure continuity and color consistency. We ablate the two loss functions to evaluate the contribution of each loss function to the whole pipeline. 
Fig.~\ref{fig:comparison_ablations} (f) and (g) demonstrate the effectiveness of the two loss functions for color reconstruction and balance for exposure correction.

\begin{table}[!t]
\begin{threeparttable}
\centering
\caption{Comparisons on computational costs of different methods.}
\label{tab:costs}
\setlength{\tabcolsep}{4pt}
\begin{tabular}{l|S[table-format=2.3,mode=text]S[table-format=3.3,mode=text]S[table-format=4.0,mode=text]S[table-format=3.2,mode=text]}
\toprule
\multirow{2}{*}{Method} &\text{Trainable} &\text{FLOPs} &\text{Inference}& \text{Inference}\\
& \text{Params (M)} &\text{(G)}& \text{Memory (MB)} & \text{Time (ms)}\\
\midrule
RVRT~\cite{liangRecurrentVideo2022}\textsuperscript{$\diamondsuit$} & 5.306 & 217.783 &5256&374.31\\
SI-HDR~\cite{leSingleImageHDR2023}  & 62.195&97.152&2758&185.07 \\
\midrule
SDSDNet~\cite{wangSeeingDynamic2021}\textsuperscript{$\diamondsuit$}&4.318&343.208&1430&349.37 \\
Retinexformer~\cite{caiRetinexformerOnestage2023}&1.606&125.367&2186&301.48 \\
GTMeanLoss~\cite{liaoGTMeanLoss2025}&1.606&125.367&2626&297.96 \\
HVI-CIDNet~\cite{yanHVINew2025} &1.976&59.344&1252&324.99 \\
\midrule
IAT~\cite{cuiYouOnly2022}&0.091 &10.794 &840&53.04 \\
DRBN-ENC~\cite{huangExposureNormalization2022} &0.612 &97.309 &1238&97.37 \\
ECLNet~\cite{huangExposureConsistencyRepresentation2022} &0.022 &18.780 &636&109.68 \\
FECNet~\cite{huangDeepFourierBased2022}&0.152 &8.823 &768&137.64 \\
LCDP~\cite{wangLocalColor2022}&0.314 &13.143 &3452&236.46 \\
LACT~\cite{baekLuminanceawareColor2023b}&2.248 &159.810 &7746&524.65 \\
MSLT++~\cite{zhou4KResolutionPhoto2024} &0.011 &0.108 &311&52.61 \\
RECNet~\cite{liuRegionAwareExposure2024}&0.039 &21.678 &816&146.77 \\
CSEC~\cite{liColorShift2024}&1.072 &16.468 &4102&571.35 \\
CoTF~\cite{liRealTimeExposure2024}&0.310 &0.981 &652&118.63 \\
ExposureSlot~\cite{jungExposureslotExposurecentric2025a}&1.905 &75.001 &1266&148.01 \\
\midrule
VECNet~\cite{liuLearningExposure2024}\textsuperscript{$\diamondsuit$}& 10.831 &534.720&3252&657.97\\
\textbf{VECNet++}\textsuperscript{$\diamondsuit$}& 6.786&216.788 & 2554 &260.16 \\
\bottomrule
\end{tabular}
\begin{tablenotes}
    \item \textsuperscript{$\diamondsuit$} denotes video inputs.
\end{tablenotes}
\end{threeparttable}
\end{table}

\textbf{Ablations on Auxiliary Architecture.}
We further improve the prior VECNet by introducing (a) optical flow matching and (b) reducing the computational costs of illumination estimation.
Ablation results are presented in Tab.~\ref{tab:ablations} \textit{Auxiliary Architectural Improvements}.
(a) Without preprocessing operations, the original unaligned multiple frames will lead to overlay artifacts in the following convolutional layers. To alleviate such artifacts, we introduce a coarse optical-flow-based pre-alignment to reduce large spatial displacements.
(b) Prior illumination estimator is implemented as part of the latent diffusion model. Such an estimator is too heavy for our low-level vision task. To reduce the computational costs in illumination map estimation, we only adopt a simple stacked convolutional block along with a depthwise separable convolution layer. It reduces the number of parameters significantly and achieves higher inference speed, as shown in Tab.~\ref{tab:costs}.

\textbf{Ablations on the MFA Module.}
We further evaluate the design of the MFA module by comparing it with two alternative reconstruction strategies, including a DCN-based variant and a pyramid-based variant.
For the DCN-based variant, we replace the FFT-guided reconstruction branch with parameter-matched convolutional layers.
For the pyramid-based variant, we adopt a three-level pyramid for coarse-to-fine reconstruction.
As shown in Tab.~\ref{tab:ablation_MFA}, the original MFA achieves the best performance among all compared variants.
These results demonstrate that FFT-guided reconstruction is critical for handling exposure-varying video frames, whereas the DCN-based alternative is less effective under diverse exposure variations.

\begin{table}[t]
    \centering
    \caption{Ablations of alternative architectures for the MFA module.}
    \label{tab:ablation_MFA}
    \begin{tabular}{lcc}
    \toprule
    Architecture & PSNR & SSIM\\
    \midrule
    Proposed MFA & 21.84 & 0.8927 \\
    DCN-based & 19.72 & 0.8722 \\
    Pyramid-based & 20.06 & 0.8829 \\
    \bottomrule
    \end{tabular}
\end{table}

\begin{table}[t]
    \centering
    \caption{Ablation study of EAF module.}
    \label{tab:ablation_eaf}
    \begin{tabular}{lccc}
    \toprule
    Setting & Params (M)&PSNR & SSIM \\
    \midrule
    \textbf{Full EAF} & 6.786&\textbf{21.84} & \textbf{0.8927} \\
    EAF w/ Zero Diff& 6.786 & 20.11 & 0.8889 \\
    EAF w/ Shuffled Diff& 6.786 & 19.47 & 0.8715 \\
    EAF w/o Attention & 6.785& 18.60 & 0.8916 \\
    Conv Replacement & 6.765& 16.29 & 0.8513 \\
    \bottomrule
\end{tabular}
\end{table}

\begin{table}[!t]
    \centering
    \caption{Ablation study on the guidance strategy in the TSR module.}
    \label{tab:ablation_TSR}
    \begin{tabular}{lcc}
    \toprule
    Guidance Strategy & PSNR & SSIM\\
    \midrule
    Latent guidance in $T_1$ + image guidance in $T_2$ & 21.84 & 0.8927 \\
    Image guidance in both $T_1$ and $T_2$ & 19.98 & 0.8391 \\
    Latent guidance in both $T_1$ and $T_2$ & 19.77 & 0.8608 \\
    Image guidance only in $T_1$ & 18.91 & 0.8036 \\
    Latent guidance only in $T_1$ & 20.17 & 0.8594 \\
    \bottomrule
    \end{tabular}
\end{table}

\textbf{Ablations on EAF Module.}
To investigate whether the performance gain stems from exposure-aware modeling, attention-based reweighting, or increased model capacity, we conduct a series of ablations on DIMEPlus.
Specifically, we replace the inter-frame differences $\tilde{F}_d$ with a zero tensor (\textit{EAF w/ Zero Diff}), randomly shuffle the inter-frame differences while preserving its marginal distribution (\textit{EAF w/ Shuffled Diff}), remove the attention mechanism (\textit{EAF w/o Attention}), and replace the EAF module with stacked convolutional layers with a comparable number of parameters (\textit{Conv Replacement}).
The results are reported in Tab.~\ref{tab:ablation_eaf}.
Replacing the inter-frame differences with zeros leads to a noticeable performance drop, while shuffling the differences further degrades the results.
It suggests that the meaningful inter-frame difference cues for estimating exposure-aware pseudo gradients are essential for the performance gain.
Furthermore, removing the attention mechanism also causes a clear degradation, indicating that adaptive feature reweighting is an important component of EAF.
Finally, the parameter-matched convolutional replacement performs consistently worse than all EAF variants, suggesting that the gain is not solely attributable to model capacity.
Overall, the results verify that both the exposure-aware pseudo gradients derived from inter-frame differences and the inverted-attention design are critical to the effectiveness of EAF.

\textbf{Ablation on the Guidance Strategy in TSR.}
We further investigate the guidance strategy in the two-stage synthesis restoration (TSR) module. The proposed TSR uses latent illumination embeddings in the first stage $T_1$ and introduces the image-level candidates $I_t^u$ and $I_t^o$ in the second stage $T_2$. This design follows the coarse-to-fine strategy of the two stages. 
The latent embeddings provide compact exposure-aware guidance for coarse restoration without imposing explicit appearance constraints.
By contrast, the image-level candidates contain more explicit appearance and illumination information, but may also include synthesis artifacts.
Therefore, we introduce them in $T_2$, where they provide explicit references for detailed refinement.
To validate this design, we compare different guidance strategies, including image guidance in both stages, latent guidance in both stages, image guidance only in $T_1$, and latent guidance only in $T_1$.
As shown in Tab.~\ref{tab:ablation_TSR}, the proposed strategy achieves the best performance. Using image guidance in the first stage leads to inferior results, suggesting that early image-level constraints may disturb coarse restoration. Using latent guidance in both stages or removing image-level guidance from $T_2$ also degrades the performance, indicating that explicit image-level candidates are beneficial for the final refinement stage. These results demonstrate that both the representation type and its stage placement are important for the effectiveness of TSR.

\begin{table*}[!t]
\centering
\caption{Quantitative analysis of the Fourier-prior validity under different motion and saturation conditions.}
\label{tab:fourier_prior_validity}
\renewcommand{\arraystretch}{1.15}
\begin{tabular}{@{}llrrrrr@{}}
\toprule
\textbf{Dataset} &
\textbf{Condition} &
\textbf{Flow p90} &
\textbf{Clip Ratio} &
\textbf{Amp. Deviation} &
\textbf{Raw Phase} &
\textbf{Learned Phase} \\
\midrule

Extreme Condition Set &
All samples &
0.0139 &
0.0059 &
1.0753 &
0.3143 &
0.5969 \\

Extreme Condition Set &
High motion / Low saturation &
0.0521 &
0.0000 &
0.8656 &
0.0403 &
0.5652 \\

Extreme Condition Set &
High motion / High saturation &
0.0230 &
0.0348 &
1.0251 &
0.0575 &
0.6277 \\

\midrule

DIMEPlus &
All samples &
0.0029 &
0.0047 &
1.1780 &
0.6268 &
0.6266 \\

DIMEPlus &
Low motion / Low saturation &
0.0002 &
0.0000 &
1.3639 &
0.8330 &
0.6393 \\

DIMEPlus &
High motion / Low saturation &
0.0092 &
0.0000 &
0.9826 &
0.3316 &
0.6142 \\

DIMEPlus &
Low motion / High saturation &
0.0002 &
0.0189 &
1.2743 &
0.8153 &
0.6253 \\

DIMEPlus &
High motion / High saturation &
0.0055 &
0.0038 &
0.9854 &
0.2620 &
0.6204 \\

\bottomrule
\end{tabular}

\vspace{1mm}
\parbox{0.98\linewidth}{
\footnotesize
\textit{Note:}
Flow p90 denotes the normalized 90th-percentile optical-flow magnitude.
The motion and saturation groups are defined using dataset-specific tertiles.
The phase values denote cosine similarity with the aligned center-frame phase.}
\end{table*}

\begin{table}[!t]
\centering
\caption{Ablation study of various input frames.}
\label{tab:ablation_frames}
\begin{tabular}{ccS[table-format=3.2,mode=text]cc}
\toprule
\multirow{2}{*}{\text{\#Frames}} & Trainable & \text{Inference} & \multirow{2}{*}{PSNR} & \multirow{2}{*}{SSIM} \\
& Params. (M) & \text{Time (ms)} & & \\
\midrule
1 & 6.678 &  70.60& 20.23&0.8819 \\
3 & 6.732 & 160.29& 19.89&0.8759 \\
5 & 6.786 & 260.16& 21.84&0.8927 \\
\bottomrule
\end{tabular}
\end{table}

\textbf{Ablations on Number of Input Frames.}
We evaluate the effect of temporal context by using 1, 3, and 5 consecutive input frames, where 5 frames correspond to our default setting.
As shown in Tab.~\ref{tab:ablation_frames}, the single-frame setting lacks temporal cues. The 3-frame setting slightly underperforms the single-frame setting, whereas the 5-frame setting achieves the best performance. This result suggests that effective temporal aggregation may require sufficient complementary observations to compensate for exposure inconsistency and residual misalignment in dynamic scenes.
In terms of complexity, increasing the number of input frames leads to marginal parameter growth but substantially higher inference cost due to additional frame-wise computation.

\begin{figure}[t]
    \centering
    \includegraphics[width=1.0\linewidth]{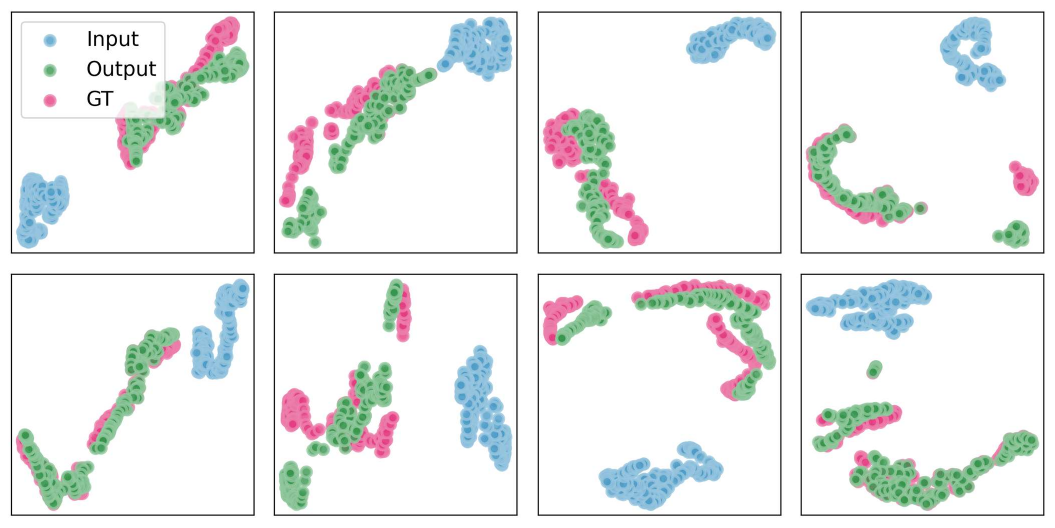}
    \caption{t-SNE visualization for part of DIMEPlus testing set. Each subplot corresponds to an individual testing sequence, showing the distribution of inputs, our outputs, and ground truths. The visualization demonstrates that the processed results align more closely with the ground truths compared to the original inputs.}
    \label{fig:visual_tSNE}
    \vspace{-1em}
\end{figure}

\textbf{t-SNE Evaluation.}
In addition, we present the statistical visualization of the results in latent space in Fig.~\ref{fig:visual_tSNE}. After being processed by our method, the underexposure and overexposure representations tend to intersect with the corresponding ground truth. 
This suggests that the proposed method effectively narrows the representation gap between ill-exposed inputs and their corresponding references.

\subsection{Validity Analysis of the Fourier Prior}
\label{subsec:fourier_prior_validity}
The Fourier prior is a conditional assumption whose reliability depends on the temporal consistency and exposure diversity of neighboring frames. 
To investigate its validity under challenging extreme conditions, we evaluate large inter-frame motion and local saturation using the collected extreme condition set and the DIMEPlus dataset.

We measure the normalized 90th-percentile optical-flow magnitude~\cite{delmerico2019we} and the persistent clipping ratio to characterize motion and saturation, respectively. We further measure the deviation of the learned aggregated amplitude from the temporal amplitude statistics, together with the cosine similarity between neighboring frame phase representations and the aligned center frame phase.

The quantitative results are reported in Tab.~\ref{tab:fourier_prior_validity}.
Under a general spatial translation, the Fourier amplitude spectrum is relatively insensitive to displacement, whereas the corresponding displacement is encoded as a phase modulation. 
Consistent with this property, the results in the ``Amp. Deviation'' and ``Raw Phase'' columns show that severe motion substantially reduces raw phase consistency, whereas the learned phase representation remains comparatively stable.
Within the range of clipping ratios evaluated, local saturation does not systematically increase the global amplitude deviation, possibly because the clipped regions occupy only a limited portion of the image. Nevertheless, more extensive or spatially widespread saturation may still compromise the reliability of amplitude aggregation.

\subsection{User Study}
\begin{figure}[t]
    \centering
    \includegraphics[width=1.0\linewidth]{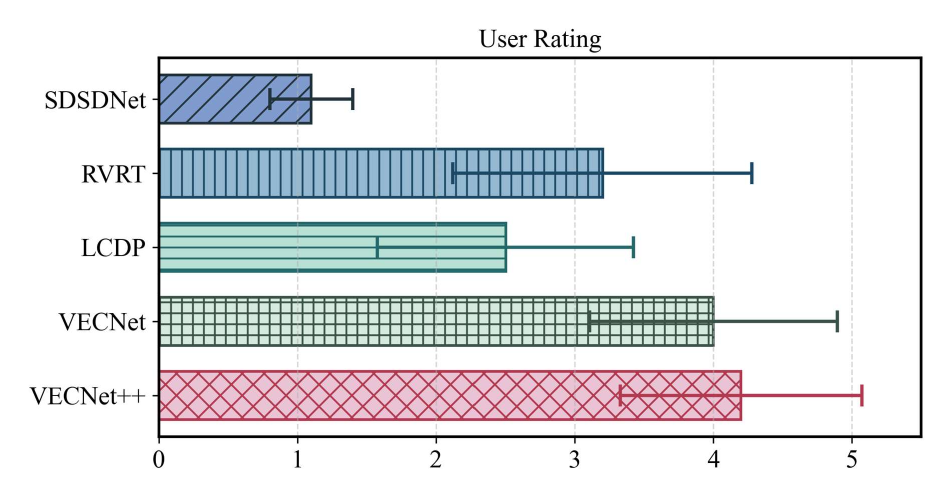}
    \caption{User study results of five methods. Participants rate the visual quality of restored videos from 1 to 5 (5 indicates the best). The y-axis lists the evaluated enhancement methods.}
    \vspace{-1em}
    \label{fig:user_study}    
\end{figure}

We conduct a user study with 20 participants to assess subjective visual quality.
Ten videos are randomly selected from the DIMEPlus dataset and enhanced using four video-based methods (SDSDNet, RVRT, VECNet, and VECNet++) and one representative image-based exposure correction method (LCDP).
For fair comparison, the input and enhanced videos are presented in random order under identical viewing conditions.
Participants rank the results from 1 (worst) to 5 (best) based on authenticity, exposure artifacts, texture contrast, color realism, and temporal stability.
The final results are summarized in Fig.~\ref{fig:user_study}.
VECNet++ achieves the highest subjective preference, demonstrating its superior perceptual quality and temporal consistency.

\section{Conclusion}
This work systematically investigates video exposure correction (VEC) from both data and method perspectives.
To address the lack of an available benchmark in this domain, we present DIMEPlus, a real-world paired video exposure correction dataset covering both underexposed and overexposed dynamic scenes with precise spatial alignment.
The dataset enables systematic training and evaluation of video exposure correction models under realistic camera motion, object motion, and exposure variations.
Building on this benchmark, we develop VECNet++, an end-to-end framework for video exposure correction. Unlike frame-wise image exposure correction, VEC requires learning a temporally stable exposure representation and reliably fusing temporal features under motion- and exposure-induced variations.
VECNet++ is formulated to address these challenges by learning an aggregated amplitude representation from consecutive frames and exploiting inter-frame discrepancies for exposure-adaptive temporal fusion. Under a Retinex-based formulation, it jointly estimates reflectance and illumination to produce temporally consistent exposure-corrected videos.
Extensive quantitative experiments and user studies demonstrate that VECNet++ consistently outperforms state-of-the-art image-based and video-based methods.
Moreover, the lightweight design significantly improves efficiency, achieving a 37\% reduction in parameters and a 60\% inference speedup compared to prior VECNet.

\begin{figure}[t]
    \centering
    \includegraphics[width=0.7\linewidth]{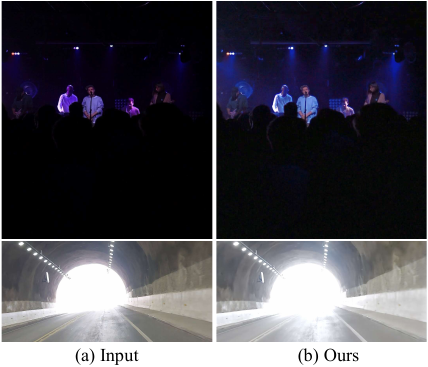}
    \caption{Limitations of our VECNet++ under extreme exposure conditions.}
    \label{fig:failure_cases}
\end{figure}

\noindent\textbf{Limitations and Future Work.}
Despite its overall effectiveness, VECNet++ still faces limitations in extremely challenging scenarios involving severe saturation and high-speed motion (Fig. \ref{fig:failure_cases}).
The Fourier prior is effective when neighboring frames preserve stable scene structures and provide complementary exposure cues.
However, severe motion disrupts cross-frame consistency, whereas saturation irreversibly removes fine-grained information from clipped regions. In these cases, the current multi-frame design may produce incomplete details or local artifacts when reliable complementary observations are unavailable.
In practical settings, addressing such issues may require additional information sources, such as multi-exposure inputs or complementary sensing modalities (e.g., event cameras).
Alternatively, generative priors could be incorporated to infer perceptually plausible structures, although pixel-level fidelity cannot be guaranteed.
In future work, we plan to explore reliability-aware integration of generative models with exposure correction frameworks to improve reconstruction quality in severely degraded images and videos while maintaining structural and temporal consistency.
\bibliographystyle{IEEEtran}
\bibliography{main}

\begin{IEEEbiography}
    [{\includegraphics[width=1in,height=1.25in,clip,keepaspectratio]{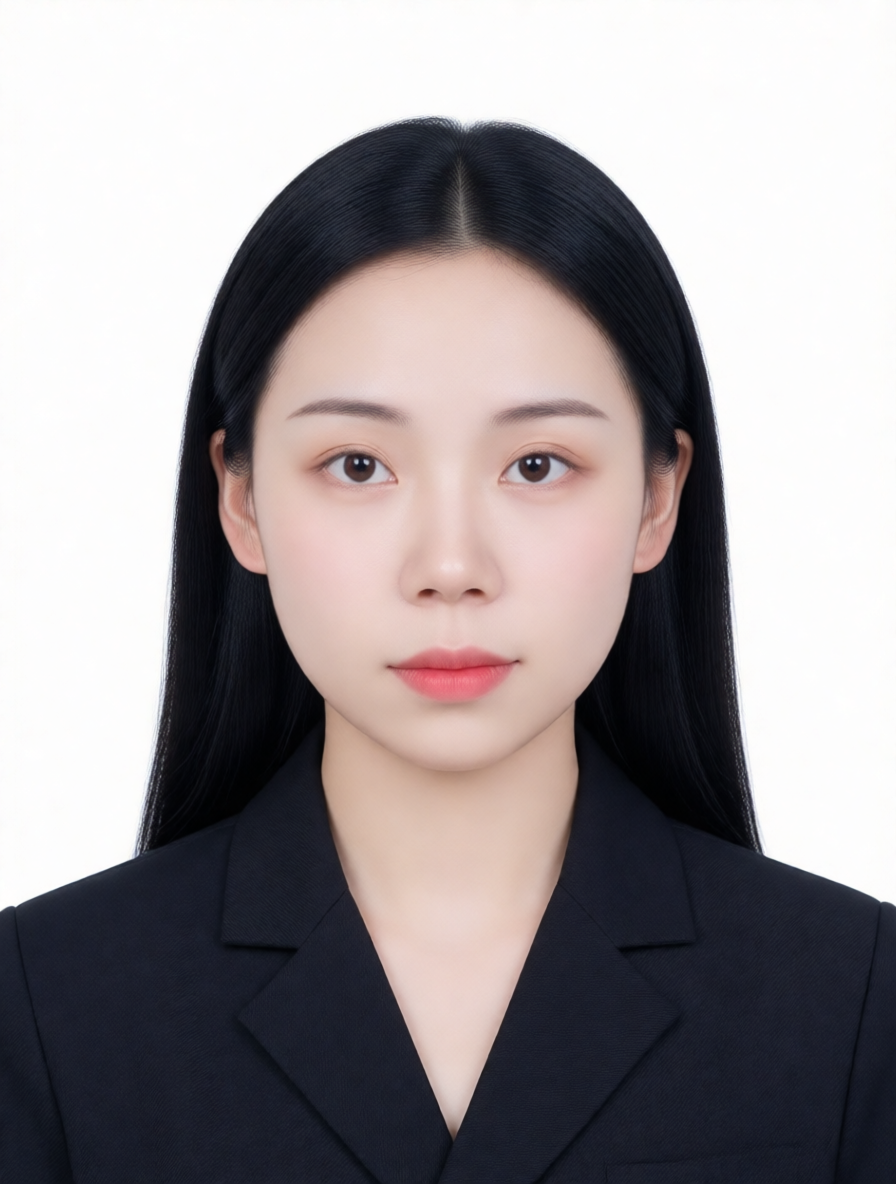}}]
    {Siru Zhang} received the B.S. degree and M.S. degree from Beijing University of Posts and Telecommunications, Beijing, China, in 2021 and 2024. She is currently pursuing a Ph.D. degree at Beijing University of Posts and Telecommunications, Beijing, China, in 2024. Her research interests include deep learning and machine learning and their applications to computer vision, especially image and video synthesis and low-quality enhancement.
\end{IEEEbiography}

\vspace{-2em}

\begin{IEEEbiography}
    [{\includegraphics[width=1in,height=1.25in,clip,keepaspectratio]{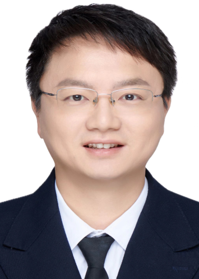}}]
    {Huiyuan Fu} (Member, IEEE) received the Ph.D. degree in computer science from the Beijing University of Posts and Telecommunications (BUPT), Beijing, China, in 2014. He is currently a Tenured Professor with the School of Computer Science, BUPT. His research interests include multimodal large language models (MLLMs), visual big data, machine learning and multimedia systems. He has authored more than 80 papers in the leading conferences and journals, including CVPR, ICCV, AAAI, ACM MM, IEEE TMM, IEEE TCSVT, IEEE TIFS, and ACM TOMM. He was the recipient of the Best Student Paper Award at IEEE ICME in 2016.
\end{IEEEbiography}

\begin{IEEEbiography}
    [{\includegraphics[width=1in,height=1.25in,clip,keepaspectratio]{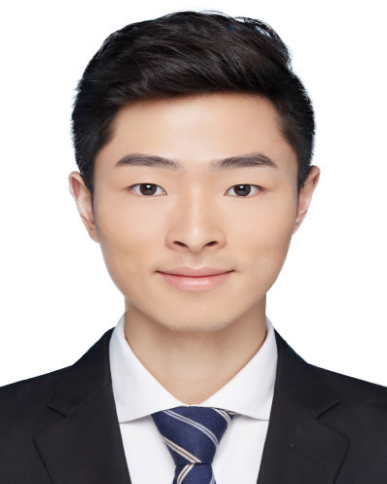}}]
    {Jin Liu} received the B.S. degree from North China Electric Power University, China, in 2020, and the Ph.D. degree from Beijing University of Posts and Telecommunications, Beijing, China, in 2025. His research interests include deep learning and machine learning, and their applications to computer vision, especially image and video synthesis and low-quality enhancement.
\end{IEEEbiography}

\begin{IEEEbiography}
    [{\includegraphics[width=1in,height=1.25in,clip,keepaspectratio]{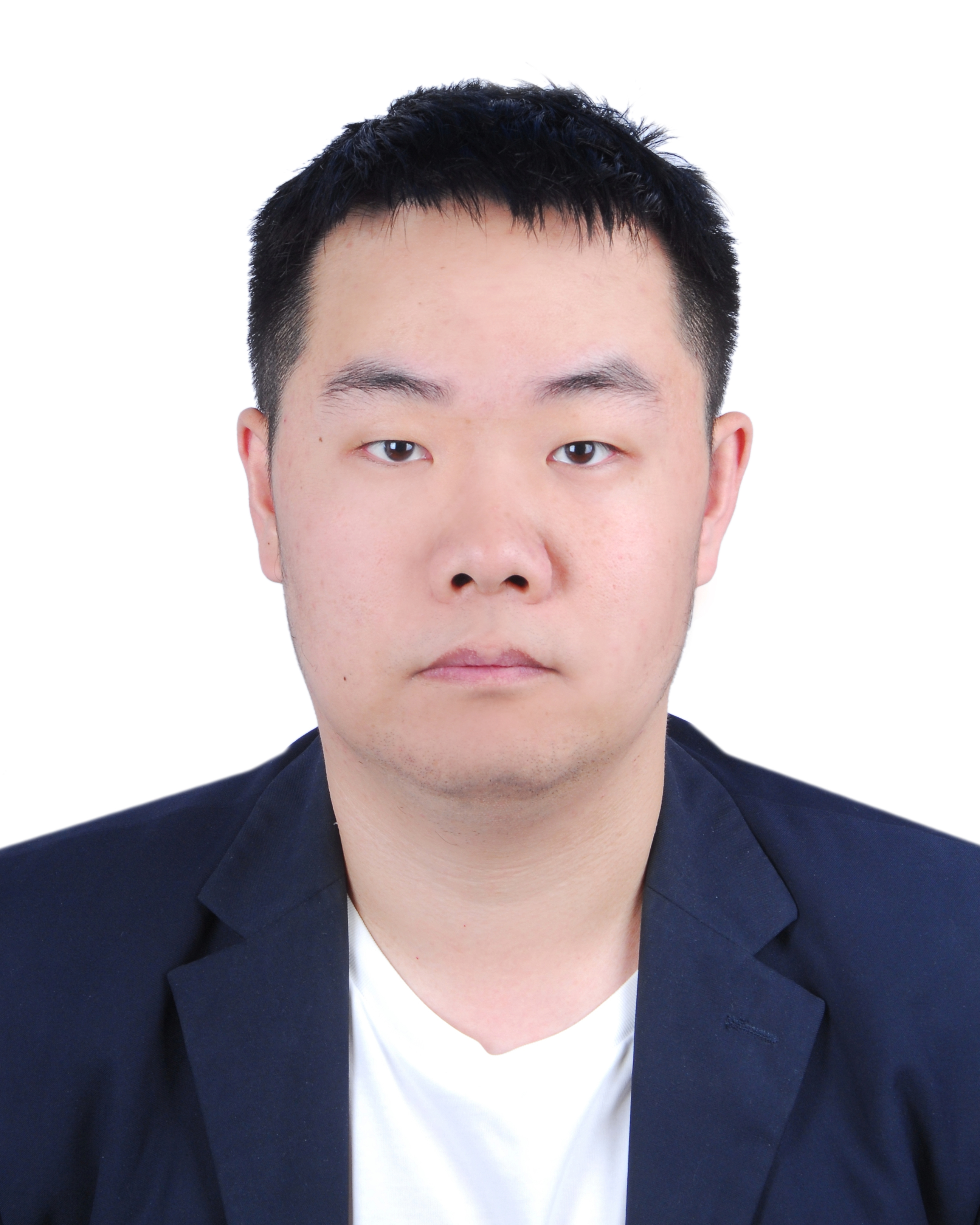}}]
    {Bo Wang} received the B.S. degree from Zhengzhou University, Henan, in 2023. He is currently pursuing the M.S. degree at Beijing University of Posts and Telecommunications, Beijing, China, in 2023. His research interests include deep learning and its applications to computer vision, especially image and video synthesis and low-quality enhancement.
\end{IEEEbiography}

\begin{IEEEbiography}
    [{\includegraphics[width=1in,height=1.25in,clip,keepaspectratio]{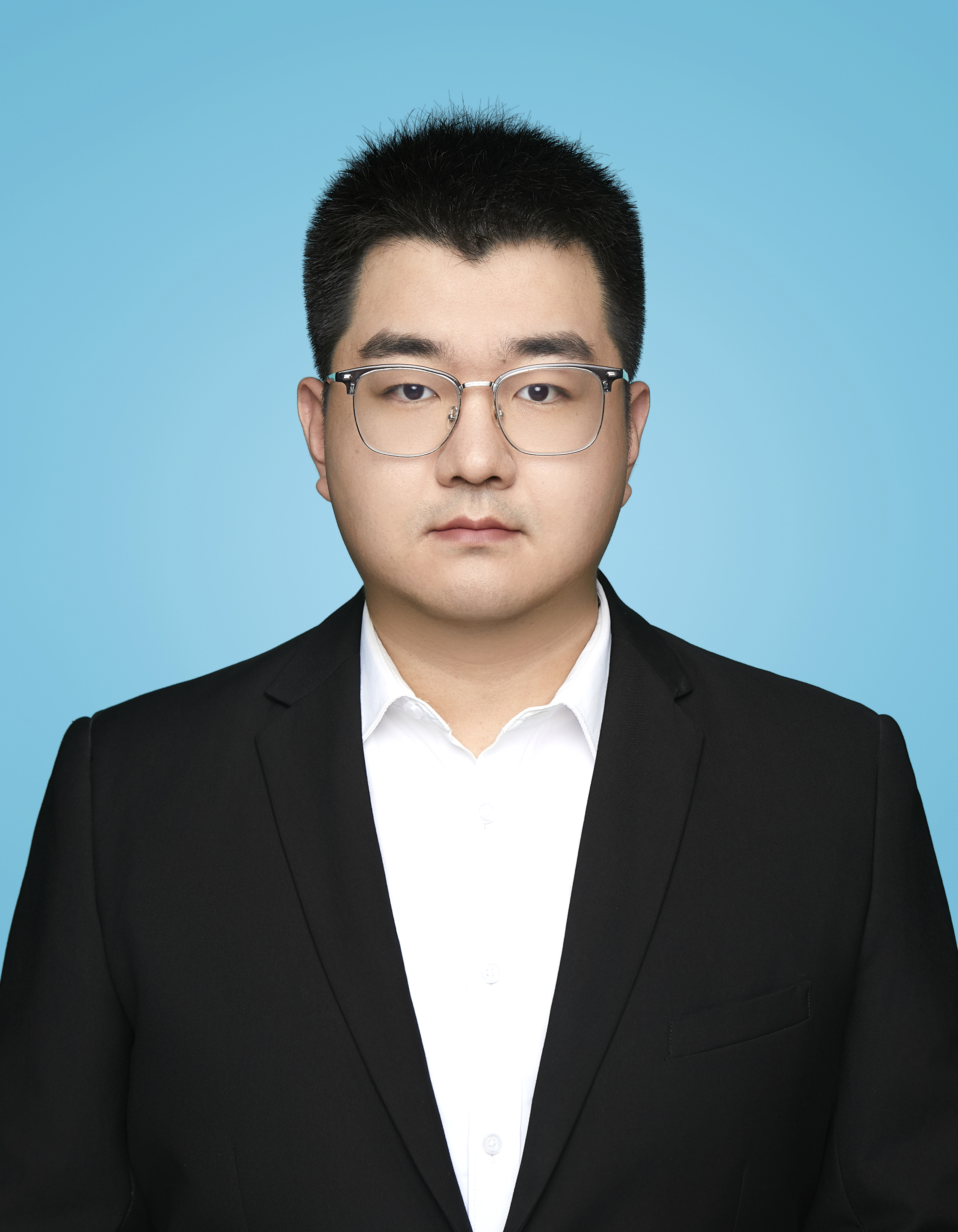}}]
    {Chuanming Wang} received the Ph.D. degree from Beijing University of Posts and Telecommunications (BUPT), Beijing, China, in 2023. He is currently a tenure-track Assistant Professor with the School of Artificial Intelligence, BUPT. His research interests include multimodal content understanding, autonomous driving, and physical AI. His work has been published in leading conferences and journals, including CVPR, AAAI, ACM MM, IEEE TMM, IEEE TCSVT, and IEEE ToN.
\end{IEEEbiography}

\begin{IEEEbiography}
    [{\includegraphics[width=1in,height=1.25in,clip,keepaspectratio]{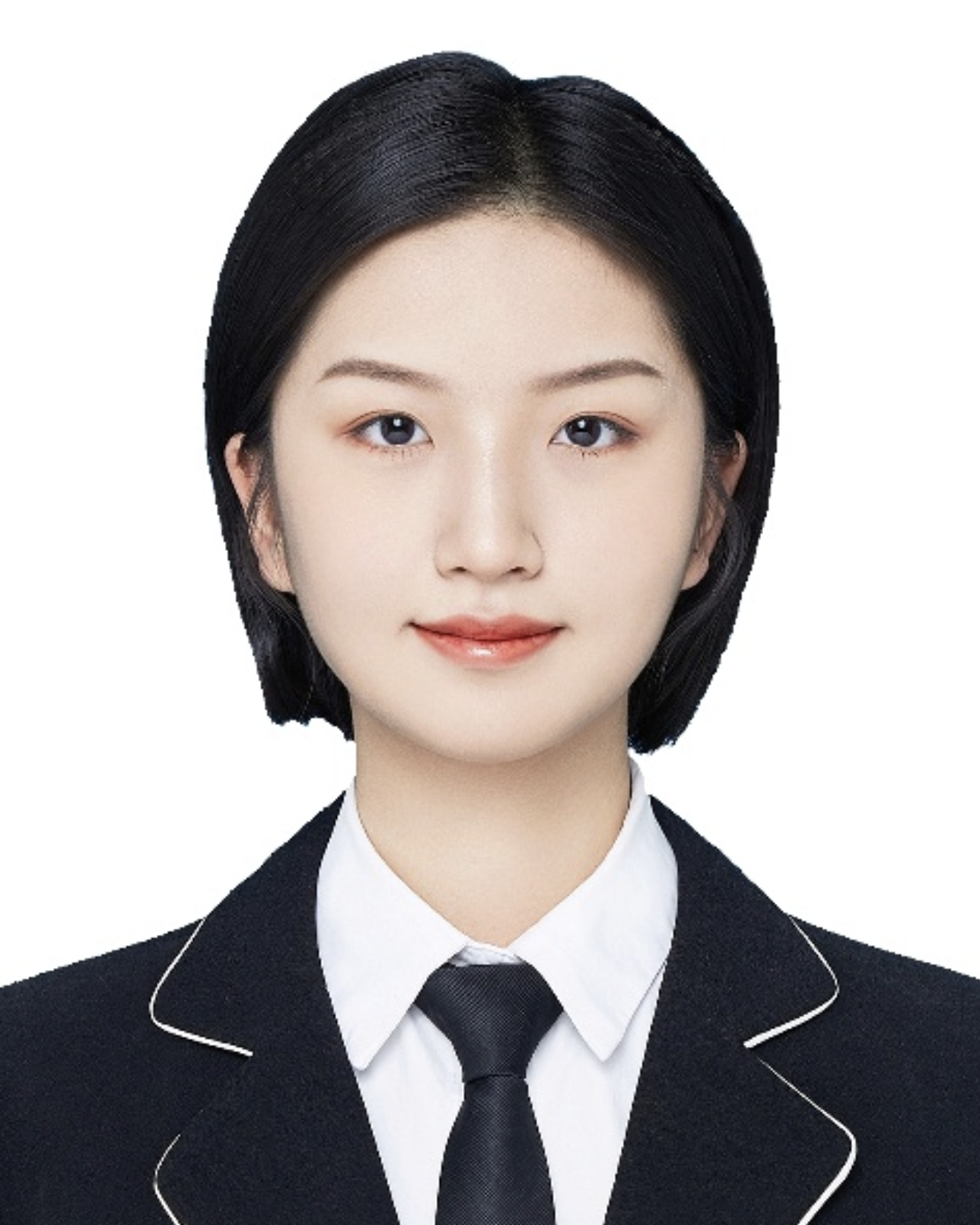}}]
    {Chongqi Sun} is currently an undergraduate student at the School of Future Technology, Beijing University of Posts and Telecommunications, Beijing, China, in 2023. Her research interests include the intersection of large foundation models and computer vision, particularly in multimodal understanding and vision-language models.
\end{IEEEbiography}
\vspace{-3\baselineskip}

\begin{IEEEbiography}
    [{\includegraphics[width=1in,height=1.25in,clip,keepaspectratio]{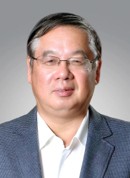}}]
    {Huadong Ma} (Fellow, IEEE) received the B.S. degree in mathematics from Henan Normal University, Xinxiang, China, in 1984, the M.S. degree in computer science from the Shenyang Institute of Computing Technology Chinese Academy of Science, China, in 1990, and the Ph.D. degree in computer science from the Institute of Computing Technology, Chinese Academy of Science, Beijing, China, 1995. He is currently a Professor of School of Computer Science, Beijing University of Posts and Telecommunications, Beijing, China. From 1999 to 2000, he held a Visiting Position with the University of Michigan, Ann Arbor, MI, USA. His current research interests include Internet of Things, sensor networks, and multimedia computing. He has authored more than 300 papers in journals, such as ACM/IEEE Transactions or conferences, such as ACM MobiCom, ACM SIGCOMM, IEEE INFOCOM, and five books. He was the recipient of the Natural Science Award of the Ministry of Education, China, in 2017, 2019 Prize Paper Award of IEEE TRANSACTIONS ON MULTIMEDIA, 2018 Best Paper Award from IEEE MULTIMEDIA, Best Paper Award in IEEE ICPADS 2010, Best Student Paper Award in IEEE ICME 2016 for his coauthored papers, and National Funds for Distinguished Young Scientists in 2009. He was/is an Editorial Board Member of the IEEE TRANSACTIONS ON MULTIMEDIA, IEEE INTERNET OF THINGS JOURNAL, and ACM Transactions on Internet of Things. He is the Chair of ACM SIGMOBILE China.
\end{IEEEbiography}

\end{document}